\documentclass[10pt,journal,compsoc]{IEEEtran}
\pdfoutput=1
\ifCLASSINFOpdf
\else
\usepackage[dvips]{graphicx}
\fi
\usepackage{url}
\usepackage{booktabs}

\usepackage[colorlinks,linkcolor=blue,anchorcolor=blue,
citecolor=blue]{hyperref}
\usepackage{graphicx}
\usepackage{amsmath,amssymb}
\usepackage{booktabs,tabularx,multirow,threeparttable}
\usepackage{float}
\usepackage{framed}
\usepackage{subfigure}
\usepackage{amssymb}
\usepackage{threeparttable}
\usepackage{multirow}
\usepackage{ragged2e}
\usepackage{makecell}
\usepackage{diagbox}
\usepackage{subfigure}
\usepackage{pifont}
\ifCLASSOPTIONcompsoc
  \usepackage[nocompress]{cite}
\else
  \usepackage{cite}
\fi
\def\ie{\emph{i.e.}}
\def\eg{\emph{e.g.}}

\def\etal{{\em et al.~}}

\newcommand{\myPara}[1]{\vspace{10pt}\noindent$\bullet$~\textbf{#1} \quad}

\newcommand{\secref}[1]{Section \ref{#1}}
\newcommand{\figref}[1]{Fig.~\ref{#1}}

\begin{document}
\title{Learning to Holistically Detect Bridges from Large-Size VHR Remote Sensing Imagery}
\author{
Yansheng Li, Junwei Luo, Yongjun Zhang, Yihua Tan, Jin-Gang Yu, Song Bai

\thanks{Yansheng Li, Junwei Luo, and Yongjun Zhang are with the School of Remote Sensing and Information Engineering, Wuhan University, Wuhan 430079, China (e-mail: yansheng.li@whu.edu.cn; luojunwei@whu.edu.cn; zhangyj@whu.edu.cn). \\
Yihua Tan is with the School of Artificial Intelligence and Automation, Huazhong University of Science and Technology, Wuhan 430074, China (e-mail: yhtan@hust.edu.cn). \\
Jin-Gang Yu is with the School of Automation Science and Engineering, South China University of Technology, Guangzhou 510641, China (e-mail: jingangyu@scut.edu.cn). \\
Song Bai is with the ByteDance AI Lab, Beijing 100098, China (e-mail: songbai.site@gmail.com).}
}
\markboth{}%
{Shell \MakeLowercase{\textit{et al.}}: Bare Demo of IEEEtran.cls for IEEE Journals}

\IEEEtitleabstractindextext{
    \justify
    \begin{abstract}
    Bridge detection in remote sensing images (RSIs) plays a crucial role in various applications, but it poses unique challenges compared to the detection of other objects. In RSIs, bridges exhibit considerable variations in terms of their spatial scales and aspect ratios. Therefore, to ensure the visibility and integrity of bridges, it is essential to perform holistic bridge detection in large-size very-high-resolution (VHR) RSIs. However, the lack of datasets with large-size VHR RSIs limits the deep learning algorithms' performance on bridge detection. Due to the limitation of GPU memory in tackling large-size images, deep learning-based object detection methods commonly adopt the cropping strategy, which inevitably results in label fragmentation and discontinuous prediction. To ameliorate the scarcity of datasets, this paper proposes a large-scale dataset named GLH-Bridge comprising 6,000 VHR RSIs sampled from diverse geographic locations across the globe. These images encompass a wide range of sizes, varying from 2,048 $\times$ 2,048 to 16,384 $\times$ 16,384 pixels, and collectively feature 59,737 bridges. These bridges span diverse backgrounds, and each of them has been manually annotated, using both an oriented bounding box (OBB) and a horizontal bounding box (HBB). Furthermore, we present an efficient network for holistic bridge detection (HBD-Net) in large-size RSIs. The HBD-Net presents a separate detector-based feature fusion (SDFF) architecture and is optimized via a shape-sensitive sample re-weighting (SSRW) strategy. The SDFF architecture performs inter-layer feature fusion (IFF) to incorporate multi-scale context in the dynamic image pyramid (DIP) of the large-size image, and the SSRW strategy is employed to ensure an equitable balance in the regression weight of bridges with various aspect ratios. Based on the proposed GLH-Bridge dataset, we establish a bridge detection benchmark including the OBB and HBB tasks, and validate the effectiveness of the proposed HBD-Net. Additionally, cross-dataset generalization experiments on two publicly available datasets illustrate the strong generalization capability of the GLH-Bridge dataset. The dataset and source code will be released at \url{https://luo-z13.github.io/GLH-Bridge-page/}.

\end{abstract}

}

\maketitle
\section{Introduction}\label{sec:introduction}

\begin{figure}[!t]
  \centering
  \includegraphics[width=\columnwidth]
  {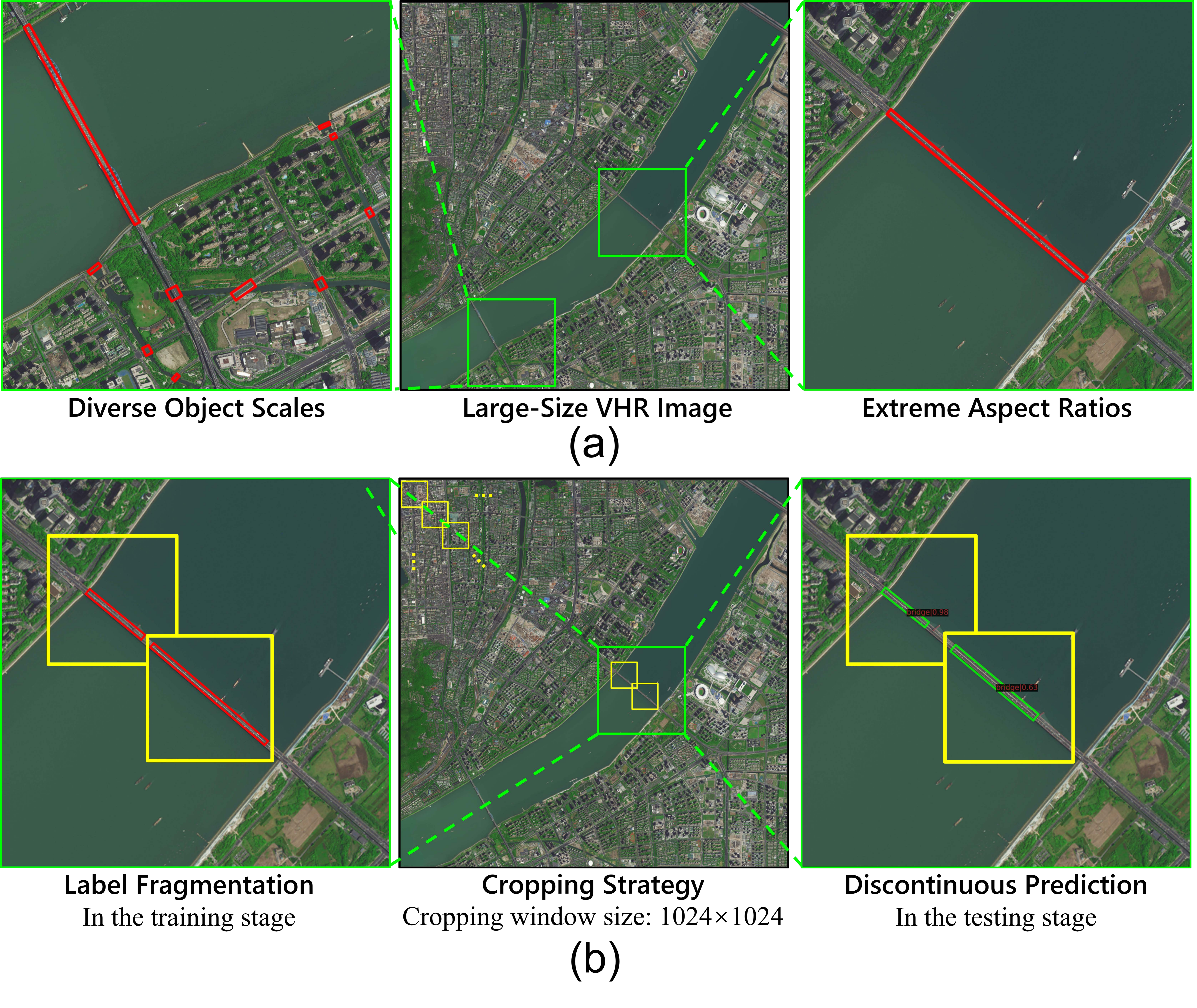}\\
  \vspace{-5pt}
  \caption{ The main characteristics of bridges impose strict requirements on both image resolution and size for bridge detection, as illustrated in (a). When tackling large-size images, the mainstream cropping strategy results in inaccurate labels and predictions. In (b), yellow windows denote the sliding windows (\ie, cropping windows), while red OBBs denote the labels and green OBBs show the prediction results.}
  \label{fig:challenges}
\end{figure}

\IEEEPARstart{B}{ridges} represent critical infrastructure components, serving as fundamental transportation facilities that traverse various landscapes. They hold substantial significance in the domains of civil transportation, military maneuvers, and disaster relief efforts~\cite{sithole2006bridge}. Meanwhile, bridges exhibit rapid construction and frequent modification. For example, in 2012, the United States had about 617,000 bridges whose deterioration will increase over the next 50 years, requiring more than \$125 billion for a backlog of repairs\footnote{https://infrastructurereportcard.org/cat-item/bridges-infrastructure}. Therefore, efficient and effective bridge detection is of paramount importance to the timely update of the navigation map and further contributes to monitoring the structural health and condition of bridges~\cite{hester2012wavelet,cantero2015bridge}. Remote Sensing Images (RSIs), with their extensive geographic coverage and high revisit frequency, are well-suited as the foundational data for bridge detection. Meanwhile, considering the powerful feature representation abilities of deep networks, deep learning-based bridge detection from RSIs holds substantial promise and has become a focal point of research\cite{guo2021accurate}.

\begin{table*}
	\centering
	\caption{Comparison between GLH-Bridge and the other relevant bridge detection datasets. Only the bridge category is selected for comparison among the multi-class object detection datasets. The comparison includes the number of images, image size, ground sampling distance (GSD), the number of instances, annotation type, backgrounds, and data source.}
        \renewcommand{\arraystretch}{1.7}
	\resizebox{1.0\textwidth}{!}{
            \large
            \begin{tabular}{c c c c c c c c}
            \hline
            Dataset & \ \thead{\large{Number}\\\large{of images}}  & Image size & GSD & \ \thead{\large{Number}\\\large{of instances}} & \thead{\large{Annotation}\\\large{type}} & \thead{\large{Diverse}\\\large{backgrounds}}   & Data source \\ \hline
            \textbf{Bridge subset of multi-class object detection dataset}\\
            NWPU VHR-10\cite{cheng2014multi} & 124 & 497$\times$693$\sim$606$\times$1,100 & 0.08$ \sim $2 & 124 & HBB &  \ding{53} & multi-source \\ 
            FAIR1M\cite{sun2022fair1m} & 581 & 1,000$\times$1,000$\sim$10,000$\times$10,000 & 0.3$\sim$0.8 & 1,008 & OBB & \ding{53} & GF, GoogleEarth  \\ 
            DOTA-v1.0\cite{xia2018dota} & 288 & 800$\times$800$\sim$4,000$\times$4,000   & 0.5 & 2,541 & OBB &  \ding{53} &  multi-source \\ 
            DOTA-v2.0\cite{ding2021object} & 382 & 800$\times$800$\sim$20,000$\times$20,000  & 0.5 & 3,043 & OBB & \ding{53} & multi-source \\ 
            DIOR-R\cite{cheng2022anchor} & 1,576 & 800$\times$800 & 0.5$\sim$30 & 4,000 & OBB & \large{\checkmark} & Google Earth  \\ 
            HRRSD\cite{zhang2019hierarchical} & 4,570 & 152$\times$152$\sim$10,569$\times$10,569  & 0.15$\sim$1.2 & 4,570 & HBB & \ding{53} & Google Earth   \\ 
            \hline
            \textbf{Dedicated bridge detection dataset}\\
            Bridges Dataset\cite{nogueira2016pointwise} & 208 & 4,800$\times$2,843 & 0.5 & 322 & HBB & \large{\checkmark} & Google Earth \\
            BridgeDetV1\cite{guo2021accurate} & 4,972 & 668$\times$668$\sim$1,000$\times$1,000  & 2$\sim$6 & 8,371 & OBB/HBB & \ding{53} & GF, GoogleEarth\\ \hline
            
            \textbf{GLH-Bridge (Ours)} & \textbf{6,000} &2,048$\times$2,048$\sim$16,384$\times$16,384 & 0.3$\sim$1 & \textbf{59,737} & OBB/HBB & \large{\checkmark} &Google Earth, Mapbox \\ \hline
            \end{tabular}
            }
	\label{table:available bridge dataset}
\end{table*}

As illustrated in \figref{fig:challenges}, detecting multi-scale bridges in RSIs is quite challenging compared to other common objects, primarily due to two main characteristics: \textbf{ (i) diverse object scales.} In VHR RSIs, the lengths of bridge instances vary from a few to several thousand pixels. \textbf{(ii) extreme aspect ratios.} There are significant differences in the degree of elongation among different bridges. To ensure the detectability of small or narrow bridges, the utilization of very-high-resolution (VHR) images is crucial. At the same time, to pursue the structural integrity of large and elongated bridges in VHR images, it is essential to conduct holistic bridge detection in large-size images, which imposes strict requirements on both datasets and methods. Despite notable advancements in multi-class object detection~\cite{xu2020gliding,yang2022detecting,nie2022multi,yang2022scrdet++} and bridge detection~\cite{nogueira2016pointwise,guo2021accurate,liu2022offshore}, there remains a deficiency in large-scale datasets and appropriate methods for holistic bridge detection in large-size VHR RSIs.

As shown in Table \ref{table:available bridge dataset}, although numerous popular datasets for object detection in RSIs have been created \cite{xia2018dota,li2020object,ding2021object,sun2022fair1m}, the quantity of bridges within these datasets is limited. Furthermore, datasets explicitly created for bridge detection \cite{nogueira2016pointwise,guo2021accurate} are often constrained by sample volumes and image sizes. Some of the existing datasets only provide horizontal bounding box (HBB) annotations instead of the accurate oriented bounding box (OBB) annotations. Therefore, training a robust and generalizable bridge detection model using the aforementioned datasets seems to be unrealistic. In response to the data limitations, we construct \textbf{GLH-Bridge}, a large-scale dataset for bridge detection in large-size VHR RSIs. GLH-Bridge contains 6,000 VHR RSIs sampled globally and over 59k manually annotated bridges. Compared with existing datasets for bridge detection, GLH-Bridge stands out by annotating multi-scale bridges in large-Size VHR RSIs that encompass various background types such as \emph{vegetation}, \emph{dry riverbeds}, and roads, thereby better capturing the characteristics of bridges in real-world scenarios. In short, the GLH-Bridge exhibits comprehensive advantages and notable merits compared with existing bridge detection datasets.

To advance the research on the fundamental and practical issue, we propose a new challenging yet meaningful task: \textbf{holistic bridge detection in large-size VHR RSIs}. To address this task, the potential solutions can be categorized into four main aspects: \textbf{(i)} Given the constraints of GPU memory, mainstream deep learning-based object detection methods \cite{ding2019learning,han2021redet,xie2021oriented,nie2022multi,yang2022scrdet++}  commonly employ cropping strategies \cite{xia2018dota,akyon2022sahi}. However, such strategies have inherent limitations and easily cut off large bridges, as shown in \figref{fig:challenges}. In addition to the cropping strategy, several object detection methods tackle the original large-size images with fixed-window downsampling strategies \cite{deng2020global,wu2021hierarchical,chen2023coupled}, resulting in a significant loss of image information; \textbf{(ii)} Methods like streaming\cite{pinckaers2020streaming} perform the forward and backward pass on smaller tiles of the large-size image, but they are unable to support deep neural network (DNN) with normalization; \textbf{(iii)} Methods like LMS\cite{le2019automatic} use memory offload to share memory across system memory (CPU DRAM) and the GPU memory. However, they introduce significant time overhead and are constrained by the maximum memory expansion rate; \textbf{(iv)} Multi-GPU tensor parallelization techniques\cite{shazeer2018mesh, xu2023efficient} have the promise to extend deep networks to support holistic processing of large-size images. However, they tend to be resource-intensive and difficult to operate in regular conditions. In summary, existing methods are ineffective under common computational resources (\eg, a single GPU with 24 GB memory) for holistic bridge detection in large-size VHR RSIs.

Considering the limitations of the aforementioned potential solutions, we propose a \textbf{h}olistic \textbf{b}ridge \textbf{d}etection \textbf{n}etwork (\textbf{HBD-Net}) specifically designed for bridge detection in large-size VHR RSIs. Our method presents two key merits: \textbf{(i)} The separate detector-based feature fusion (SDFF) architecture, when applied to the dynamic image pyramid (DIP), demonstrates an efficient approach for processing large-size images with minimal resource consumption. \textbf{(ii)} The shape-sensitive sample re-weighting (SSRW) strategy balances regression weights of bridges with different aspect ratios. Experimental results on GLH-Bridge demonstrate the outstanding performance of our proposed HBD-Net.

To sum up, this paper makes the first exploration of holistic bridge detection in large-size VHR RSIs as far as we know. The main contributions of this paper are summarized as follows:

\begin{itemize}
\item We propose GLH-Bridge, the first large-scale dataset for bridge detection in large-size VHR RSIs. With 59,737 bridges set against various backgrounds, this dataset offers a comprehensive representation of bridges in real-world scenarios.

\item A cost-saving network for holistic bridge detection in large-size images (i.e., HBD-Net) is proposed, which can efficiently handle large-size images with the common GPU and holistically detect multi-scale bridges with the well-designed SDFF architecture and SSRW strategy.

\item  Using the proposed GLH-Bridge dataset, we create a benchmark for bridge detection, covering both the OBB and HBB tasks. The HBD-Net achieves superior performance compared to existing state-of-the-art algorithms. Furthermore, we conduct cross-dataset generalization experiments to demonstrate the strong generalization ability of GLH-Bridge. We hope this benchmark can contribute to the fundamental evaluation of object detection in large-size images.

\end{itemize}

The rest of this paper is organized as follows: Section 2 provides an overview of existing datasets and algorithms for bridge detection. Section 3 offers a detailed description of the proposed GLH-Bridge dataset. In Section 4, we introduce the proposed HBD-Net. Section 5 presents the experimental results. Finally, Section 6 concludes the paper and provides insights for future work.

\section{Related Work}\label{sec:related}
In this section, we first discuss available datasets for bridge detection. Next, we briefly review bridge detection methods and potential methods from relevant fields for object detection in large-size images.

\subsection{Datasets for Bridge Detection in Remote Sensing Images}
As shown in Table \ref{table:available bridge dataset}, the existing datasets available for bridge detection can be categorized into multi-class datasets that incorporate the bridge category and specialized datasets explicitly designed for bridge detection.

\subsubsection{ Multi-Class Datasets for Bridge Detection}
In the literature, numerous large-scale and high-quality remote sensing object detection datasets have been proposed. For example, NWPU VHR-10\cite{cheng2014multi} is a dataset with ten categories, expanding the category of geospatial objects. DOTA\cite{xia2018dota} and DIOR\cite{li2020object} have raised the number of instances to a new level, reflecting the prevalence of multi-class objects in remote sensing scenes. FAIR1M\cite{sun2022fair1m} accomplishes a more detailed classification taxonomy of geospatial objects. Despite these datasets containing the bridge category, they have limited quantities of bridge instances. As summarized in Table \ref{table:available bridge dataset}, these multi-class datasets are unable to fulfill the aforementioned three criteria of an ideal bridge detection dataset: \textbf{large volume of samples}, \textbf{large-size image}, and \textbf{VHR image}.

It is noted that the existing multi-class object detection benchmarks\cite{ding2021object,li2020object} show that bridge is one of the most difficult categories to detect. For example, in DOTA-v1.0 and DOTA-v1.5 \cite{xia2018dota}, the highest accuracies for the bridge category in the OBB task are 64.5$\%$ and 59.6$\%$, respectively, which are obviously lower than the other classes\footnote{https://captain-whu.github.io/DOTA/results.html}. Bridge detection, particularly in the OBB task, undoubtedly poses significant challenges. Therefore, addressing the shortage of large-scale bridge detection datasets is crucial to training high-performance bridge detection models.
 
\subsubsection{Specialized Datasets for Bridge Detection}
Besides multi-class object detection datasets of aerial images, researchers have developed diverse remote sensing datasets for one specific category to facilitate more adaptable and crucial single-class object detection. As shown in Table \ref{table:available bridge dataset}, there exist two publicly available datasets \cite{nogueira2016pointwise,guo2021accurate}, which are specifically designed for bridge detection in RSIs. 

\textbf{Bridges Dataset\cite{nogueira2016pointwise}:} Keiller \etal proposed the first dataset for bridge detection and identification in VHR RSIs, known as Bridges Dataset. This dataset comprises 500 images with a consistent size of 4,800 $\times$ 2,843 pixels. The image in this dataset has a spatial resolution of 0.5m, aligning with the VHR criteria in remote sensing scenarios. It is sampled globally using ArcGIS\footnote{https://www.arcgis.com/} and annotated bridges across different types of background terrains. However, the dataset has certain limitations. It is constrained by the relatively low number of instances and offers coarse HBB annotations for the bridges. Furthermore, the bridges are primarily located at the center of the image in this dataset, which may potentially distort the learning process for bridge detection models by prior biases.

\textbf{BridgeDetV1\cite{guo2021accurate}:} Guo \etal constructed a bridge detection dataset named BridgeDetV1 for detecting waterborne bridges in RSIs. The dataset consists of 5,000 images with the spatial resolution ranging from 2 $\sim$ 6 meters and image size ranging from 668 $\times$ 668 $\sim$ 1,000 $\times$ 1,000 pixels. It encompasses a total of 8,371 bridges annotated with both HBB and OBB. Although BridgeDetV1 contains a larger number of bridges compared to previous datasets, its limited spatial resolution restricts its ability to detect small bridges. Furthermore, BridgeDetV1 only focuses on waterborne bridges, resulting in a lack of scene diversity. 

As a whole, existing dedicated datasets for bridge detection are insufficient to reflect the characteristics of bridges in real-world scenarios. Therefore, it is urgent to build a comprehensive, large-scale bridge detection dataset with large-size VHR images and rich instance types.

\subsection{Bridge Detection in Large-Size Remote Sensing Imagery}
To motivate holistic bridge detection in large-size images, we discuss methods for bridge detection in RSIs and potential technologies to cope with object detection in large-size images in the following sections.

\begin{figure*}[t!]
        \centering
            \includegraphics[width=\textwidth]{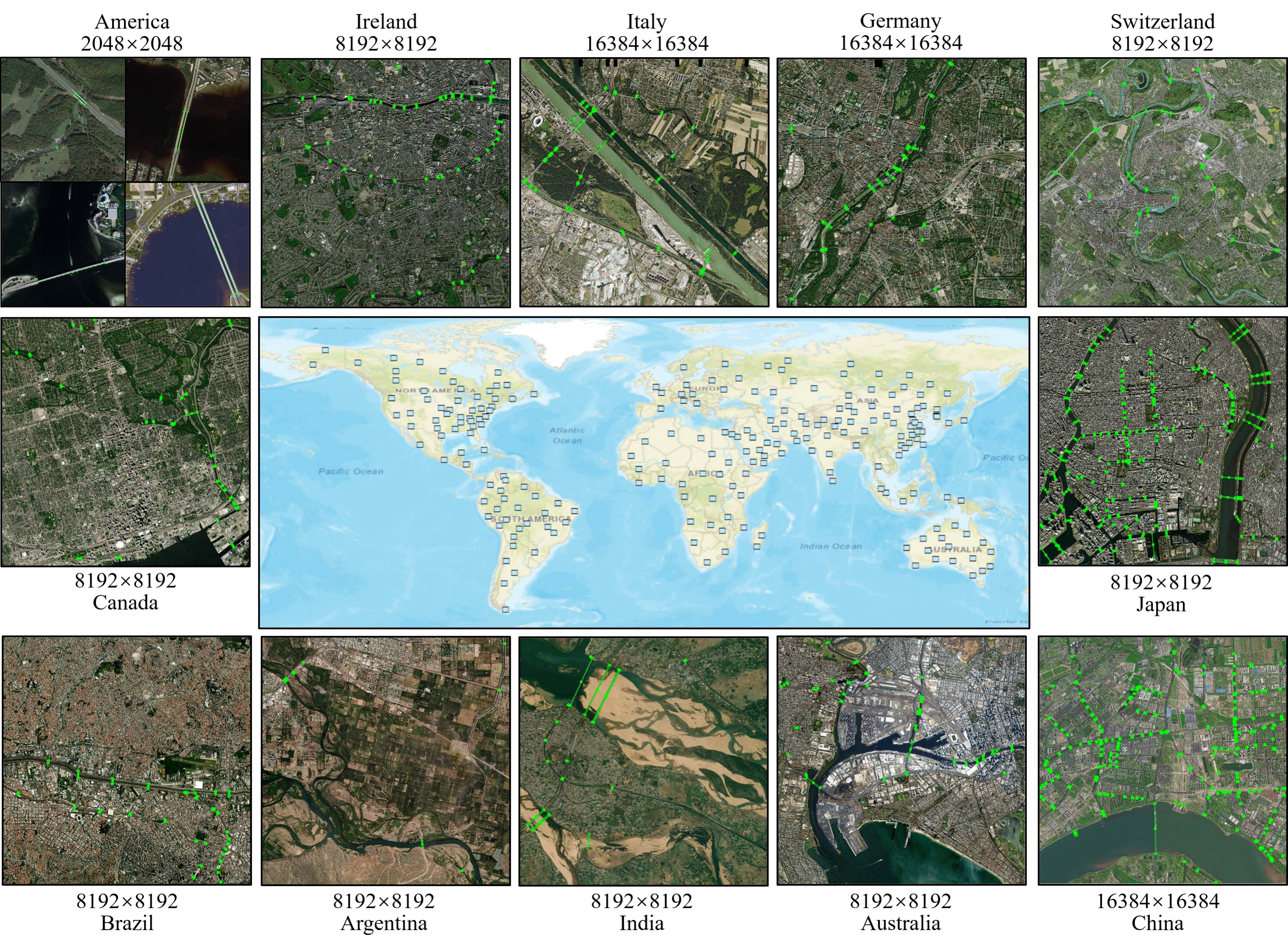}\\
           \vspace{-8pt}
	\caption{The geographical distribution map of the sampled images from the proposed GLH-Bridge dataset.}
	\label{fig:overview}
\end{figure*}

\subsubsection{Bridge Detection in Remote Sensing Imagery}
Bridge detection in RSIs is a longstanding research topic. Chaudhuri \etal \cite{chaudhuri2008automatic} utilized traditional supervised classification techniques and prior knowledge to detect bridges from multi-spectral images. Sithole \etal \cite{sithole2006bridge} focused on bridge detection in airborne scenes by detecting the cross-sectional contours of bridges. Several traditional algorithms were also developed to detect bridges in synthetic aperture radar (SAR) images based on edge and geometric features of bridges or water bodies \cite{biao2001segmentation,fulin2003algorithm,bai2005optimal}. Generally, these methods mainly relied on hand-crafted features by exploiting the bridges' geometry structure and the context of the surrounding water bodies.

Recently, some deep learning-based methods for bridge detection in RSIs have been proposed. Chen \etal \cite{chen2020new} incorporated attention modules to perform waterborne bridge detection. Guo \etal \cite{guo2021accurate} introduced the prior information of water bodies and combined bridge detection with the auxiliary task of water body segmentation. Wang \etal \cite{wang2021prior} designed a module for injecting water prior information into the bridge detection task through binary segmentation maps. Some other researchers \cite{liu2022offshore} used multi-feature fusion methods to perform bridge detection. However, these methods primarily concentrated on detecting bridges in small-size or low-resolution images. It is noted that the existing methods disproportionately prioritized water features for locating bridges, even though bridges span across diverse terrains. This overemphasis on water body information has caused biases in feature learning and failed to present practical scenarios. Hence, generalized bridge detection algorithms are still much underexplored.

\subsubsection{Object Detection in Large-Size Imagery}
In this section, we introduce methods designed for object detection in large-size images and methods borrowed from related fields that may have potential applications in tackling large-size images. It is worth noting that in large-size VHR images, object detection is more challenging than other tasks like semantic segmentation~\cite{chen2019collaborative,shan2021uhrsnet,guo2022isdnet,zheng2023farseg++} or style transfer~\cite{chen2022towards}, as the latter focuses on pixel-level details, while the former operates at the instance level.

In the field of object detection, cropping strategies like \cite{akyon2022sahi} are commonly used to handle large-size images in popular benchmarks\cite{xia2018dota,sun2022fair1m}. However, the use of the cropping strategy poses a significant risk of cutting off large bridges. As a consequence, this can lead to misalignment of the supervision signal and loss of contextual information. Moreover, some approaches have been proposed to detect objects in large-size images by downsampling the original image using a fixed size or resolution. Chen \etal \cite{chen2023coupled} proposed a coupled global-local object detection network with two branches inspired by global-local networks for segmentation \cite{chen2019collaborative}. Deng \etal \cite{deng2020global} utilized a global-local self-adaptive network to conduct drone-view object detection in large-size images via downsampling and self-adaptive cropping. However, as mentioned in \secref{sec:introduction}, such methods are not suitable for handling large-size images and can easily result in significant information loss.

Some potential deep learning-based technologies~\cite{tellez2019neural,pinckaers2021detection} can be found in the literature to handle large-size images. Pinckaers \etal proposed streaming \cite{pinckaers2020streaming}, which constructs the later activations by streaming the input image through the CNN in a tiled fashion, but it is unable to support DNN with normalization despite the fact that the normalization is a critical dependency in modern DNNs. Le \etal \cite{le2019automatic} proposed an approach based on formal rules for graph rewriting, which is able to automatically manage GPU memory to save memory usage. However, it often consumes significant computational time and is restricted by the maximum memory expansion. Additionally, Shazeer \etal \cite{shazeer2018mesh} proposed Mesh-TensorFlow for distributed tensor computations and data parallelism to address the memory problem (\eg, memory limitation of GPU). Nevertheless, Mesh-TensorFlow usually requires extensive computing resources, making them unfriendly for deployment on edge-computing devices. As a whole, it is not straightforward to extend the aforementioned methods to address holistic bridge detection in large-size RSIs. 

Hence, it is essential to develop a cost-saving approach for bridge detection that efficiently handles large-size VHR images with common GPU hardware.

\section{GLH-Bridge Dataset}\label{Sec:dataset}
Our goals for developing a new dataset for bridge detection are twofold: (i) to occupy the niche of large-scale datasets for bridge detection in large-size VHR RSIs. (ii) to promote a new meaningful yet challenging task: holistic bridge detection in large-size VHR RSIs. This section provides a comprehensive overview of the GLH-Bridge dataset, focusing on three key aspects: data collection, data annotation, and data analysis.

\subsection{Data Collection}
Taking the variations in imaging perspectives of RSIs into account and to increase data diversity, we collect images from multiple satellite sensor platforms such as Google Earth and MapBox. The GLH-Bridge dataset provides global coverage through the collection of 6,000 optical RSIs obtained from over 400 cities or regions covering Asia, Africa, South America, North America, Europe, and Oceania. The images are collected from 2019 to 2022, with the image size ranging from 2,048 × 2,048 pixels to 16,384 × 16,384 pixels, and spatial resolution varying from 0.3m to 1.0m. The overall distribution and some samples from the dataset are illustrated in \figref{fig:overview}.

To comprehensively acquire RSIs containing bridges on a global scale, we employ two distinct approaches to select candidate areas for image download. The first approach entails acquiring meta bridge information from the National Bridge Inventory (NBI)\footnote{https://www.fhwa.dot.gov/bridge/nbi.cfm}, an extensive database curated by the Federal Highway Administration. The NBI includes comprehensive details on bridges throughout the United States, including various types such as {\em highway, railway, waterborne bridges}, and {\em tunnels}. After acquiring the raw data, we filter the data on the basis of the construction years by the original database to exclude excessively outdated bridges. Subsequently, we infer the type of bridges in terms of the objects \emph{"on"} or \emph{"under"} them, to further filter out occluded bridges. Finally, we utilize Google Earth for image download. To ensure data randomness and prevent the concentration of all bridges in the center of the images, we define random windows based on the geographic coordinates during the process of download. Through the above process, we can ensure the diversity of the collected bridge types and backgrounds, fully reflecting the appearance characteristics of the bridge. To ensure that the sampling area is globally and as evenly distributed as possible, the other approach involves selecting candidate geographic areas using electronic maps and satellite images from a global range, excluding the United States. During the selection of areas of interest, we prioritize sampling in large urban areas. We first used information on the location of major cities and important rivers in each country to generate a list of cities with a high density of potential bridges. From this list, a random sample is drawn from a fixed-size area within the geographic region in each city. Subsequently, random sampling is conducted in rural areas and small towns, which are also added to the candidate geographic areas. Finally, RSIs are downloaded in terms of candidate geographic areas that exhibit diverse terrain styles and bridge backgrounds across different regions. This approach involves collecting negative samples from areas with fewer bridges, such as rural areas, islands, and deserts, in order to maintain consistency in the geographic distribution and dataset diversity. 

With the purpose of leveraging the complementary geographic coverage via the aforementioned two image collection approaches, we partition the overall dataset randomly into training, validation, and testing sets with a ratio of 6:2:2. More specifically, the training, validation, and testing sets consist of 3613, 1194, and 1193 large-size images, respectively.

\subsection{Data Annotation}
\subsubsection{Annotation Criteria}
The geographical entity \emph{"bridge"} is defined by considering both the bridge structure and its spatial context. In this vein, our visual interpretation process adheres to a stringent differentiation between bridges and roads. When dealing with suspended roads that cast shadows, we determine the two endpoints of one bridge based on the observation of whether they intersect distinct topographic features like valleys, rivers, or vegetation or not. This approach is crucial to ensure the exactitude of bridge labeling, with specific emphasis placed on the verification of objects that are susceptible to ambiguity, such as overpasses lacking topographical intersections or roads traversing regions between rice paddies.

\begin{figure}[!tb]
	\begin{center}
		\includegraphics[width=\columnwidth]{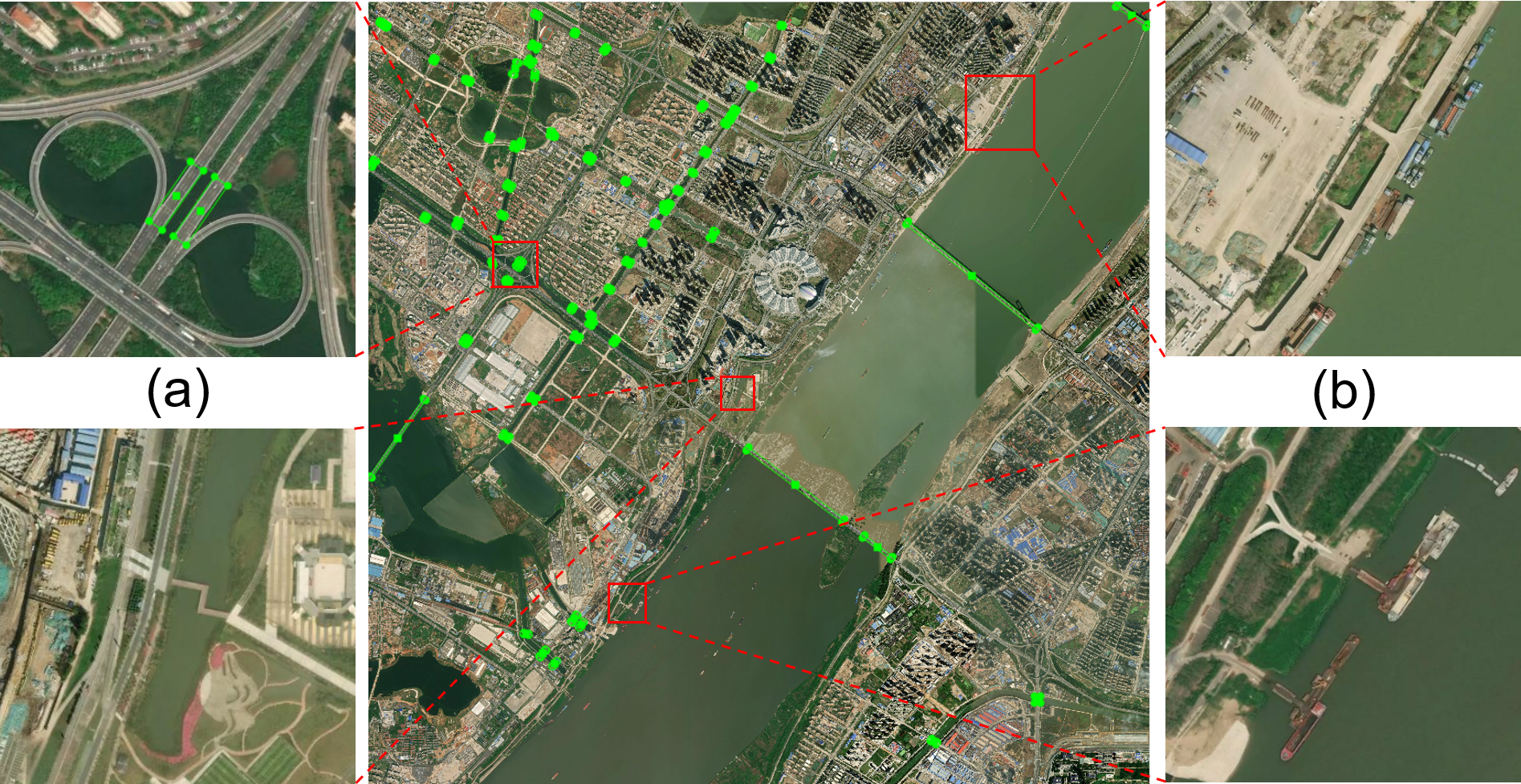}
	\end{center}
        \caption{Examples of labeling according to the criteria. (a) Roads across water with excessive curvature or an irregular shape are not labeled. (b) Two terminal connections are not labeled.}
	\label{fig:label_rule}
\end{figure}

The application of labeling criteria is illustrated in \figref{fig:label_rule}. Objects deemed to be non-bridges or bridges presenting challenges in labeling are deliberately omitted from the labeling process, such as two terminal connections shown in \figref{fig:label_rule}(a). \figref{fig:label_rule}(b) shows a road across the water with excessive curvature or an irregular shape that will not be labeled. In the process of annotating bridges, we establish the length threshold as \textbf{12 pixels} according to the size of \emph{extremely Small} in \cite{cheng2023towards}, whereby bridges shorter than this threshold will not be labeled. It should be mentioned that this approach incorporates bridges with a width less than the length threshold into the dataset, thereby introducing a notable challenge in the detection of diminutive instances.

\begin{figure*}[t!]
        \centering
            \includegraphics[width=\textwidth]{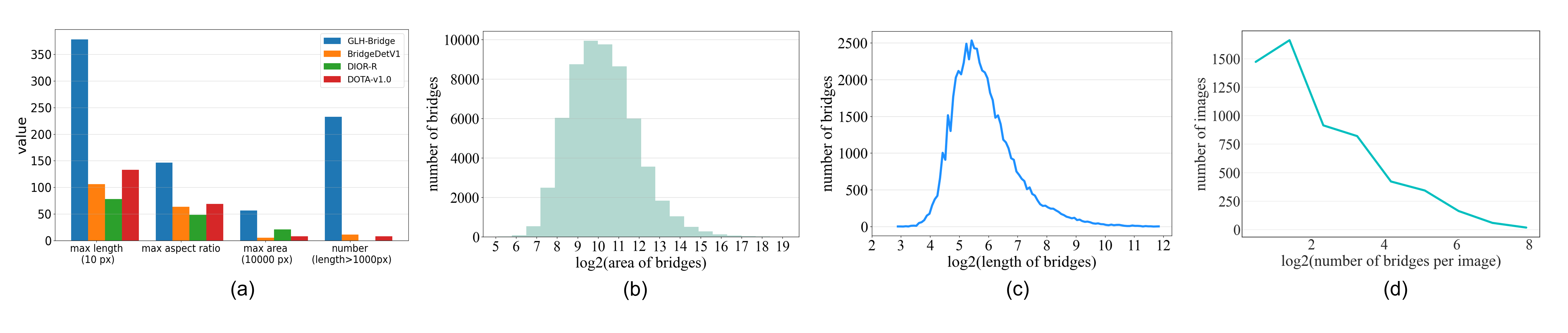}\\
           \vspace{-8pt}
	\caption{Illustration of GLH-bridge's characteristics. (a) Comparison of bridges' characteristics across different datasets. (b) Distribution of bridges' areas in GLH-Bridge. (c) Distribution of bridges' length in GLH-Bridge. (d) Distribution of bridges' density in GLH-Bridge.}
 
\label{fig:compare_max}
\end{figure*}

\subsubsection{Annotation Management}
The procedure of labeling GLH-Bridge encompasses a tripartite framework consisting of three stages: \textbf{pre-annotation stage}, \textbf{expert feedback and refinement stage}, and  \textbf{large-scale detailed annotation stage}. In light of the overhead perspective characteristic of remote sensing images, it is acknowledged that HBB is inherently limited in the ability to precisely delineate the actual positions of objects with arbitrary directions, as it contains a significant amount of irrelevant information from the background. Therefore, we use RoLabelImg\footnote{https://github.com/cgvict/roLabelImg} to manually generate the fine OBB for bridges. Specifically, the labeled rectangular bounding box can be defined by four corner points $\left (  x_{1},y_{1},x_{2},y_{2},x_{3},y_{3},x_{4},y_{4}\right )$ in the clockwise order. In the initial phase of pre-annotation, we form a specialized team comprising 10 members, each possessing extensive expertise in the field of remote sensing interpretation, is assembled. This team undergoes comprehensive training in fundamental annotation techniques and subsequently conducts annotation tests on a representative subset of the dataset. In the following feedback and refinement stage, experts thoroughly review and evaluate the team's initial annotations, resulting in the formulation of refined annotation criteria. Subsequently, guided by this adjustment, the team embark on the formal large-scale annotation process, accompanied by experts' random sampling inspections.

\begin{figure}[!htb]
  \centering
  \includegraphics[width=\columnwidth]{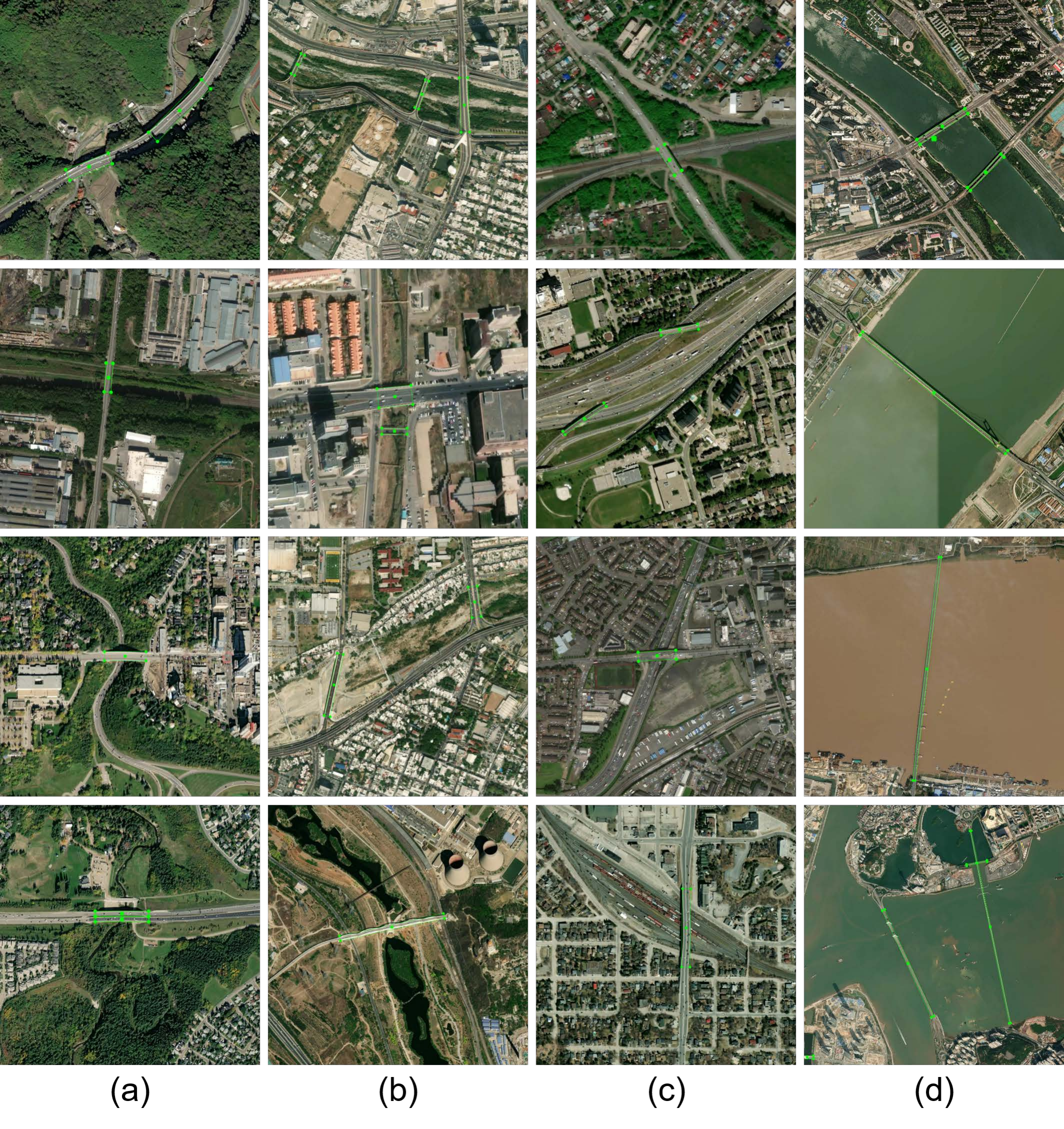}\\
  \caption{Illustration of bridges across different backgrounds in the proposed GLH-Bridge dataset. (a) Bridges across vegetation. (b) Bridges across dry riverbeds. (c) Bridges across roads. (d) Bridges across water bodies.}
  \label{fig:multi_background}
\end{figure}

\subsection{Dataset Analysis}
In contrast to the other existing bridge detection datasets, GLH-Bridge exhibits notable advantages in terms of GSD, image size, instance quantity, and instance diversity. The GLH-Bridge dataset showcases six prominent merits.

\myPara{Various Instance Scales.} 
GLH-Bridge incorporates a diverse range of bridge sizes, ranging from tiny bridges with 12 pixels to giant bridges exceeding 3000 pixels. As depicted in \figref{fig:compare_max}(a), large bridges show a high presence in GLH-Bridge, surpassing the quantity reported in existing datasets. This highlights the imperative of utilizing raw large-size images to preserve the integrity of bridges. Furthermore, as illustrated in \figref{fig:compare_max}(b) and \figref{fig:compare_max}(c), a substantial number of small bridges are showcased in GLH-Bridge. Consequently, detecting huge bridges entails processing raw large-size images, presenting a challenge in the context of conventional practice that employs small-size images for the detection of petite bridge instances.

\myPara{Extreme Aspect Ratios.}
GLH-Bridge contains many giant bridges with extreme aspect ratios, as depicted in \figref{fig:compare_max}(a). The identification of these instances poses a formidable challenge for oriented object detection algorithms. 

\myPara{Large Image Sizes.}
In the context of the GLH-Bridge, over 1,000 large-size VHR images have sizes greater than 8,000 $\times$ 8,000 pixels. Due to the diverse sizes of these images, conventional downsampling techniques using fixed ratios are ill-suited. The effective processing of these large-size images, while simultaneously preserving the integrity of exceptionally large bridges, presents a significant challenge for existing object detection methods.

\myPara{Diverse Background Types.}
As shown in \figref{fig:multi_background}, GLH-Bridge includes bridges across diverse terrains, encompassing not only \emph{water body} but also \emph{dry riverbeds}, \emph{vegetation}, \emph{valleys}, \emph{deserts}, \emph{urban roads}, \emph{etc}. This requires object detection algorithms to possess the capability to recognize bridges across a spectrum of backgrounds. Additionally, the challenge is further exacerbated by the potential for bridges to intersect or overlap with other objects, such as roads.

\myPara{Global Coverage.}
GLH-Bridge spans the globe and includes samples from all continents. This vast and diverse region provides a wide range of bridge types and landscapes, promoting the dataset's generalizability to various scenarios.

\begin{figure*}[!tb]
	\begin{center}
		\includegraphics[width=1.0\linewidth]{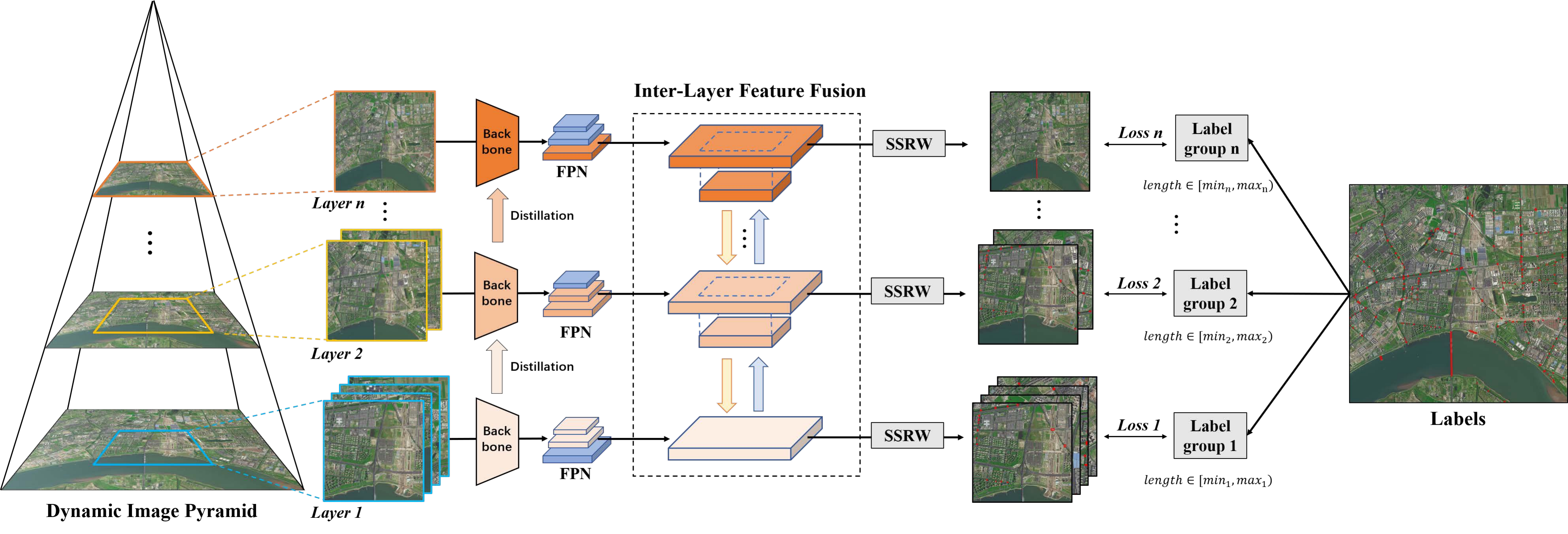}
	\end{center}
	\caption{The pipeline of the proposed HBD-Net. It contains the proposed SDFF architecture and SSRW strategy. The SDFF architecture consists of separate detectors and the IFF module. From the input large-size VHR image, we construct a DIP and send it to the separate detectors of the SDFF to obtain features. Then features from all detectors of the SDFF are fused via the IFF module to share both contextual and detailed texture information. The SSRW strategy is applied in the sample selection stage of object detectors to balance the regression weight. Finally, the output fusion features are fed into the object detectors' heads to obtain the results of each layer, which are used to compute the loss with corresponding ground-truth labels.}

    \label{fig:method}
\end{figure*}

\myPara{Variation in Instance Density.}
The distribution of bridges per image in GLH-Bridge is illustrated in \figref{fig:compare_max}(d). In densely populated urban areas or regions abundant in waterways and transportation, bridges are frequently densely distributed. However, rural areas or less developed regions exhibit a smaller number of bridges, with background areas occupying a significant proportion. 

\section{The Proposed Method}\label{sec:method}
To holistically detect bridges from large-size VHR images, this paper presents HBD-Net, which stands as the pioneering approach expressly tailored for this objective. This section is dedicated to providing a detailed explanation of the HBD-Net.

\subsection{Model Preview}
Contemporary deep networks encounter limitations when directly processing large-size RSIs due to the constrained memory capacity of the GPU. To address this problem, we present one factorized representation (i.e., the DIP) of the original large-size image. What's more, we leverage the proposed SDFF with separate detectors to train or infer upon the DIP, and an inter-layer feature fusion (IFF) module is proposed to facilitate feature complementation between layers within the SDFF. 
Moreover, we enhance the performance of HBD-Net by incorporating the SSRW strategy during sample allocation, and via cross-scale-transfer distillation in SDFF. Our method is illustrated in \figref{fig:method}.

\subsection{HBD-Net Architecture}
To effectively process the large-size image, we propose the SDFF architecture, which utilizes separate detectors to tackle the DIP and conducts feature fusion via the IFF module. We will provide a detailed explanation of these components in the following sections.

\subsubsection{Separate Detectors on Dynamic Image Pyramid}
\textbf{DIP Construction.} When presented with a large-size VHR image with a size of $H\times W$, we progressively downsample the original large-size VHR images at a fixed ratio of $\sigma$ to construct the image-level pyramid with a variable number of layers. The termination condition of the top layer (the $n$-th layer) of the pyramid is defined as follows:

\begin{equation}
\frac{H}{\sigma^{n-1}}\le H_t \quad \text{or} \quad  \frac{W}{\sigma^{n-1}}\le W_t,
\end{equation}
where $(H_t,W_t)$ is the termination threshold. So we can get the DIP with $n$ layers and the size of its top layer image is $(H_n, W_n)$, where $H_n =H/\sigma^{n-1}$, $W_n =W/\sigma^{n-1}$. At each layer of the DIP, we employ a fixed-sized window (the size is equal to $(H_t, W_t)$) to gradually extract the image patches and send them into the detector corresponding to the layer. 

\textbf{Separate Detectors.} It is noted that retaining extremely small labels in the downsampled layers can lead to severe information loss. Additionally, the identification of tiny objects in layers with higher resolution tends to be more accurate. Against this backdrop, a set of thresholds is introduced to allocate the OBB labels to each layer of the DIP based on the OBB's length. As a result, each detector embedded within the SDFF is responsible for predicting bridges with specific scales. To enable the SDFF to possess scale sensitivity when detecting multi-scale bridges, we utilize separate object detectors at layers of the SDFF instead of a unified detector (the reason is explained in \ref{sec:DIPFF}). Overall, one large-size VHR image is decomposed into one DIP with multiple layers, which passes through the SDFF followed by separate detectors. Therefore, this factorized framework enables the training of HBD-Net even when computational resources are limited (\eg, one single GPU).

\subsubsection{Inter-Layer Feature Fusion}

Considering the varying field-of-views in the same window at different layers of DIP, the higher layers have global information, while the lower layers contain detailed information. To effectively utilize complementary cues to feature fusion, we devise an Inter-Layer Feature Fusion (IFF) module to enable bidirectional feature sharing within the SDFF.

\begin{figure}[!htb]
  \centering
  \includegraphics[width=\columnwidth]
  {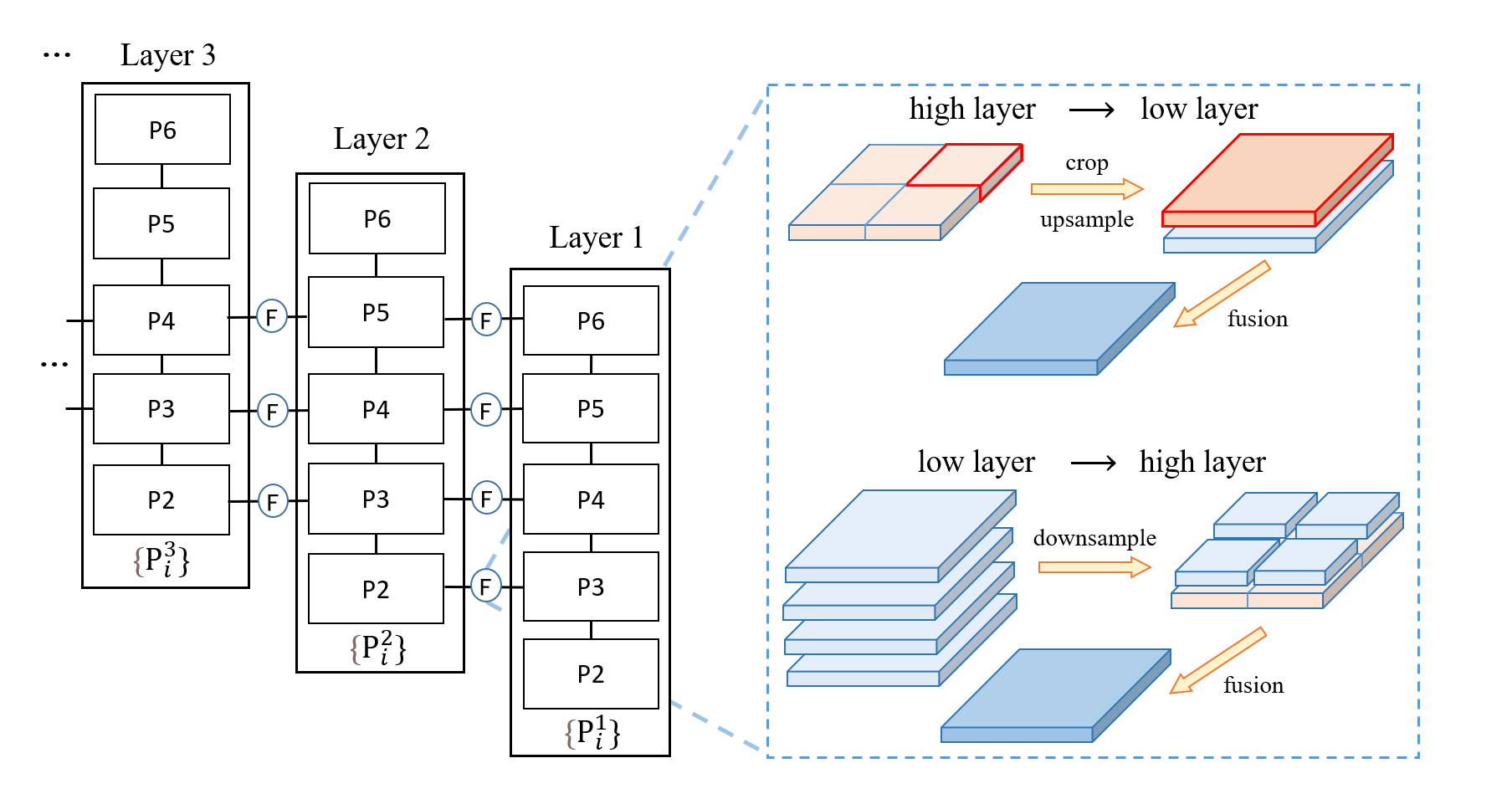}\\
  \caption{Illustration of the proposed IFF module. The figure illustrates the ways of feature fusion between two adjacent layers.}
  \label{fig:DIPFF}
\end{figure}

Similar to the basic feature extractors (\eg, Resnet\cite{he2016resnet} followed by FPN\cite{lin2017FPN}), this paper recommends extracting feature pyramids from the DIP. Given the feature sets obtained by the feature pyramid network (FPN) from all image layers within the DIP, candidate feature sets are selected from the adjacent image layers. Subsequently, we perform inter-layer feature fusion on these candidate feature sets via feature alignment and fusion.

\textbf{Feature Selection.} We begin by identifying candidate feature sets for fusion. As shown in \figref{fig:DIPFF}, in the case of FPN, the spatial sizes of adjacent levels in the feature pyramid always differ by $2\times $. We set $P_i^{j}$ as the $i$-th level feature in the feature pyramid of the $j$-th image layer. Assuming that the $j$-th image layer is located in the middle of the DIP, the feature pyramids of the two neighboring layers can be represented as $P^{j-1}$ and $P^{j+1}$, respectively. Given that the downsampling ratio of FPN equals the downsampling ratio $\sigma$ we set for DIP, we can draw the conclusion that the following features $P_{i+1}^{j-1}$, $P_{i}^{j}$, and $P_{i-1}^{j+1}$ have the same actual downsampling ratio. We define this candidate feature set as $P_{cand} = \{{P_{i+1}^{j-1}, P_{i}^{j}, P_{i-1}^{j+1}\}}$. 

\textbf{Feature Alignment and Fusion.} After obtaining the candidate feature sets $P_{cand}$, we align the features within $P_{cand}$ based on consistent spatial position and conduct fusion. The rough representation of this process is shown in \figref{fig:DIPFF}. For $P_{i+1}^{j-1}$, we begin by conducting downsampling it and then align it with the spatial consistent region on $P_{i}^{j}$. For $P_{i-1}^{j+1}$, we align it with the spatial consistent region on $P_{i}^{j}$ by cropping. Due to the existence of sliding windows, the image features of the $(j-1)$-th layer are extracted in batches during the training process. Consequently, these features are concatenated along the spatial dimension to fit the size of $P_{i}^{j}$ after downsampling. Following this alignment process, we perform feature fusion as follows:
\begin{equation}
P_{i}^{j} = act(conv(\text{\emph{concat}}(\text{\emph{align}}( \{ {P_{i+1}^{j-1}, P_{i}^{j}, P_{i-1}^{j+1} \}} )))),
\end{equation}
where $act$, $conv$, $\text{\emph{concat}}$ and $\text{\emph{align}}$ refer to activation layer (\eg, sigmoid), $1\times 1$ convolutional layer, channel-wise concatenation operation and aforementioned alignment process, respectively. The feature $P_{i}^{j}$ will be updated after fusion. For the remaining features that are not possible to construct $P_{cand}$, we preserve the original feature without performing feature fusion. After the IFF module, the fused features are obtained for subsequent detection tasks. In this way, the features in each layer of the SDFF are fused with the features from the adjacent layers, allowing the feature in the middle layer to capture contextual information from the upper layer and detailed texture information from the lower layer.

\begin{figure}[!htb]
  \centering
  \includegraphics[width=\columnwidth]
  {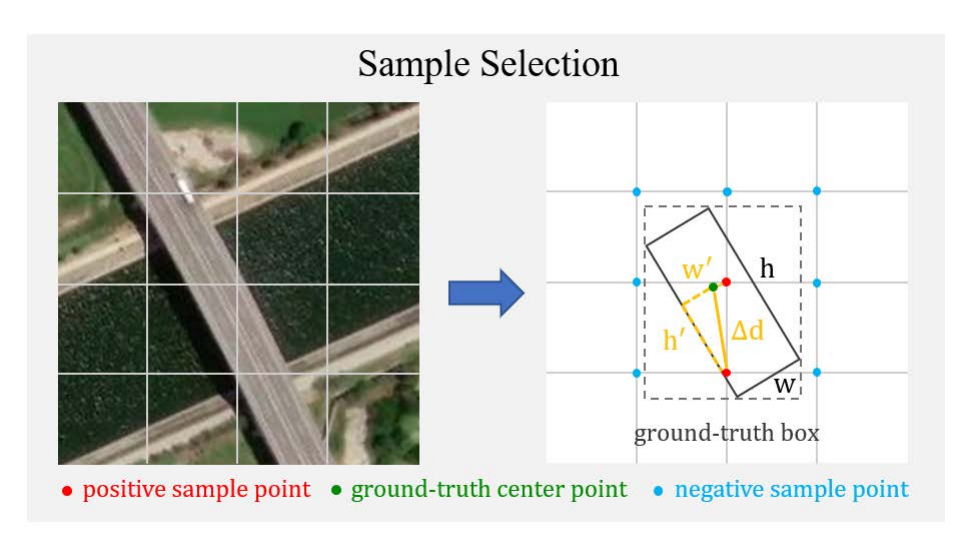}\\
  \caption{Illustration of the proposed SSRW strategy. The red and blue points represent positive and negative samples selected by the object detector, respectively. For anchor-based detectors, these points correspond to the feature map locations generating anchors or proposals. For anchor-free detectors, these points indicate the grids on the feature maps. To maintain clarity and simplicity, the depiction of anchors or proposals associated with the sample points (applicable to anchor-based methods) is not depicted in this illustration.}
  \label{fig:pointweight}
\end{figure}

\subsection{HBD-Net Optimization}

As bridges exhibit drastic variations in spatial scales and aspect ratios, it is essential to acknowledge that the Intersection over Union (IoU) between the prediction and label exhibits heightened sensitivity to regressive bias, especially for ground-truth boxes with larger aspect ratios. Existing oriented object detection methods usually employ fixed strategies like the max intersection over union (IoU)\cite{ding2019learning,xie2021oriented} or distance in feature maps\cite{tian2019fcos} to select positive samples and give all samples the same weight. However, this practice is unsuitable, as it fails to account for the disparities in regression weights required for samples with distinct aspect ratios. To address this problem, we propose a shape-sensitive sample re-weighting (SSRW) strategy during the sample assignment stage. It aims to encourage the deep network to prioritize samples with extreme aspect ratios, and further balance the weighted regression losses.

As illustrated in \figref{fig:pointweight}, following the assignment and selection of positive and negative samples (\ie, sample points), each ground-truth box is linked to its positive samples for subsequent regression and classification predictions. For the positive samples corresponding to a ground-truth box, where $w$ and $h$ represent the width and height of the ground-truth box, respectively, and $r$ denotes the normalized aspect ratio of ground-truth boxes within the mini-batch. The distance between the center point of this box and one of its corresponding positive samples is denoted as $\Delta d$. From this, the projected lengths $w^\prime$ and $h^\prime$ of $\Delta d$ in the $w$ and $h$ directions can be computed. The relative offset factors $r_w$ and $r_h$ are then defined as $r_w = \frac{2w'}{w}$, $r_h = \frac{2h'}{h}$. After acquiring the relative offset factors, we use offset measurement factors $Q^w$ and $Q^h$ to evaluate the deviation of the selected samples. These factors can be expressed as:
\begin{align}
Q^w &= \ln(r_w+1)+1, \\
Q^h &= \ln(r_h+1)+1.
\end{align}

After obtaining the offset measurement factors $Q^w$ and $Q^h$, the SSRW strategy incorporates them into the regression loss weight $w^{reg}$ to assign higher weights to more challenging samples (\ie, those with larger aspect ratios). The $w^{reg}$ is defined as follows:
\begin{equation}
w^{reg}=\mu Q^w Q^hr,
\end{equation}
where $\mu$ is the adjustment factor. In this case, an increased value of $Q^w$ and $Q^h$ indicates a larger relative distance between the positive sample's prediction box and the ground-truth box. This suggests that the transformation of the candidate box into a high-quality regression box is more challenging. Consequently, assigning a higher $w^{reg}$ to such positive samples enables the detector to prioritize them. Additionally, due to the predominance of small objects, when the $w^{reg}$ shift towards objects with larger aspect ratios, it contributes to achieving an equitable balance in regression weights among bridges with varying aspect ratios.

The total loss of oriented object detection and horizontal object detection is defined as follows:
\begin{align}
\mathcal{L_{O}}&=\sum_{m=1}^{n} \lambda ^m(\frac{1}{N}\sum_{i\in \psi}\mathcal{L}_{i}^{cls} + \frac{1}{N^+}\sum_{j\in \psi _p} w^{reg}_{j}\mathcal{L}_{j}^{reg}), \\
\mathcal{L_{H}}&=\sum_{m=1}^{n} \lambda ^m(\frac{1}{N}\sum_{i\in \psi}\mathcal{L}_{i}^{cls} + \frac{1}{N^+}\sum_{j\in \psi _p} \mathcal{L}_{j}^{reg}),
\end{align}
where $\small{w^{reg}_{j}}$ is the regression weight calculated by the proposed SSRW strategy. $n$ is the number of layers in the DIP, and $\lambda ^m$ is the balanced weight corresponding to the loss of the $m$-th layer, which is set to 1. $\psi$ and $\psi_p$ represent the set of all samples and the set of positive samples, respectively. $N$ and $N^+$ denote the total number of all samples and positive samples, respectively. The classification loss $\mathcal{L}_{i}^{cls}$ is focal loss\cite{lin2017focal} and the regression loss $\mathcal{L}_{j}^{reg}$ is Smooth L1 loss as defined in \cite{girshick2015fast}.

To make full use of the supervision of multi-scale bridges and pursue the scale-sensitive detector, we train the separate detectors within the SDFF layer-by-layer. This process commences with training the bottom layer and proceeds to train each subsequent layer, culminating with the top layer. Upon the completion of training for the detector of the $m$-th layer ($m\in [1,n-1]$ ), we use its weights to initialize the detector of the $(m+1)$-th layer. Meanwhile, congenetic labels from the label assign strategy are used to constrain the scale-equivalence of outputs from separate detectors, achieving cross-scale-transfer distillation and enhancing the performance of the deep network.

\section{Experimental Results and Analysis}
\label{sec:experiment}

\subsection{Benchmark}

\begin{table*}[tb!]
    \centering
    \caption{Accuracy ($\%$) of OBB and HBB tasks on GLH-Bridge. * indicates training the HBD-Net without the proposed SSRW strategy. \dag~indicates training the GLSAN without the Local Super-Resolution Network (LSRN).}
    \renewcommand{\arraystretch}{1.2} 
    \setlength\tabcolsep{2.6mm}
    \resizebox{1.0\textwidth}{!}{
    \begin{tabular}{l|c|c c c|c c c c}
    \hline
        \textbf{OBB Task} & Backbone & $\mathrm{mAP}$ & $\mathrm{AP_{50}}$ & $\mathrm{AP_{75}}$ & $\mathrm{AP_{sh}}$ & $\mathrm{AP_{md}}$ & $\mathrm{AP_{lg}}$ & $\mathrm{AP_{hg}}$  \\ \hline
    
        Faster R-CNN-O\cite{xia2018dota} & R50-FPN & 31.35 & 67.99 & 22.73 & 30.54  & 35.08  & 18.52  & 5.41 \\
        RoI-Transformer\cite{ding2019learning} & R50-FPN & 33.66 & 69.58 & 25.55 & 32.28  & 38.05  & 26.32  & 3.10  \\ 
        FCOS-O\cite{tian2019fcos} & R50-FPN & 29.28 & 60.14 & 22.74 & 26.98  & 33.78  & 25.02  & 2.13 \\ 
        R$^3$Det\cite{yang2021r3det} & R50-FPN & 31.11 & 68.01 & 22.84 & 29.95  & 34.04  & 23.11  & 4.70 \\ 
        KLD\cite{yang2021learning} (R$^3$Det) & R50-FPN & 31.92 & 68.47 & 23.67 & 30.88  & 35.15  & 23.32  & 5.83   \\ 
        ReDet\cite{han2021redet} & ReR50-ReFPN & 34.29 & 69.99 & 26.06 & 31.94  & 38.12  & 29.49  & 2.47   \\ 
        Oriented R-CNN\cite{xie2021oriented} & R50-FPN & 34.16 & 69.87 & 26.29 & 32.83  & 37.74  & 29.30  & 5.68   \\ 
        Oriented RepPoints\cite{li2022oriented} & R50-FPN & 29.66 & 60.19 & 22.73 & 26.65  & 34.27  & 19.09  & 7.44   \\ 
        CGL \cite{chen2023coupled} & R50-FPN  &34.72 & 70.55 & 27.47 & 33.16 & 38.14 & 30.53 & 12.68 \\
       HBD-Net (Ours) & R50-FPN & \textbf{35.35} & \textbf{71.69}  & \textbf{28.69}  & \textbf{33.38}  & \textbf{38.93}  & \textbf{33.47}  & \textbf{20.61} \\ \hline
       \textbf{HBB Task} & Backbone & $\mathrm{mAP}$ & $\mathrm{AP_{50}}$ & $\mathrm{AP_{75}}$ & $\mathrm{AP_{sh}}$ & $\mathrm{AP_{md}}$ & $\mathrm{AP_{lg}}$ & $\mathrm{AP_{hg}}$  \\ \hline
               Faster R-CNN\cite{ren2015faster} & R50-FPN  & 33.40          & 70.72          & 30.73          & 31.63          & 40.14          & 30.49          & 8.19           \\
RetinaNet\cite{lin2017focal}     & R50-FPN  & 30.71          & 67.30          & 27.32          & 28.96          & 37.02          & 24.59          & 3.39           \\
FCOS\cite{tian2019fcos} & R50-FPN & 22.32          & 51.33          & 18.01           & 19.42          & 27.99          & 15.05          & 3.41           \\
TOOD\cite{feng2021tood}& R50-FPN & 30.43          & 65.01            & 28.52          & 28.04          & 37.05             & 26.41          & 6.41           \\
Cascade R-CNN\cite{cai2018cascade}  &R50-FPN & 33.71          & 70.84          & 32.10            & 31.84          & 39.50           & 32.09          & 8.01           \\
ATSS\cite{zhang2020bridging}&R50-FPN  & 27.92          & 63.52          & 23.00            & 27.16          & 33.44          & 19.44          & 5.91           \\
GuidingAnchor\cite{wang2019region}& R50-FPN  & 33.81          & 71.22          & 31.71          & 31.72          & 39.89          & 31.54          & 7.11           \\ 
GLSAN\dag~ \cite{deng2020global} & R50-FPN  &26.95 & 54.51 & 18.13 & 22.65  & 31.24 & 21.24 & 13.29 \\
SAHI \cite{akyon2022sahi} & R50-FPN  & 34.00 & 71.12 & 30.94 & \textbf{32.73}  & 41.11  & 31.51  & 18.68  \\
CGL \cite{chen2023coupled} & R50-FPN  &33.93 & 71.25 & 30.47 & 31.91 & 40.28 & 32.62 & 16.12 \\
HBD-Net* (Ours)  & R50-FPN & \textbf{34.49} & \textbf{72.45} & \textbf{32.68} & 32.29 & \textbf{41.54} & \textbf{35.21} & \textbf{35.59} \\ \hline
       
        \end{tabular}}
	\label{tab:obbres}
\end{table*}

\begin{figure*}[t!]
        \centering
            \includegraphics[width=\textwidth]{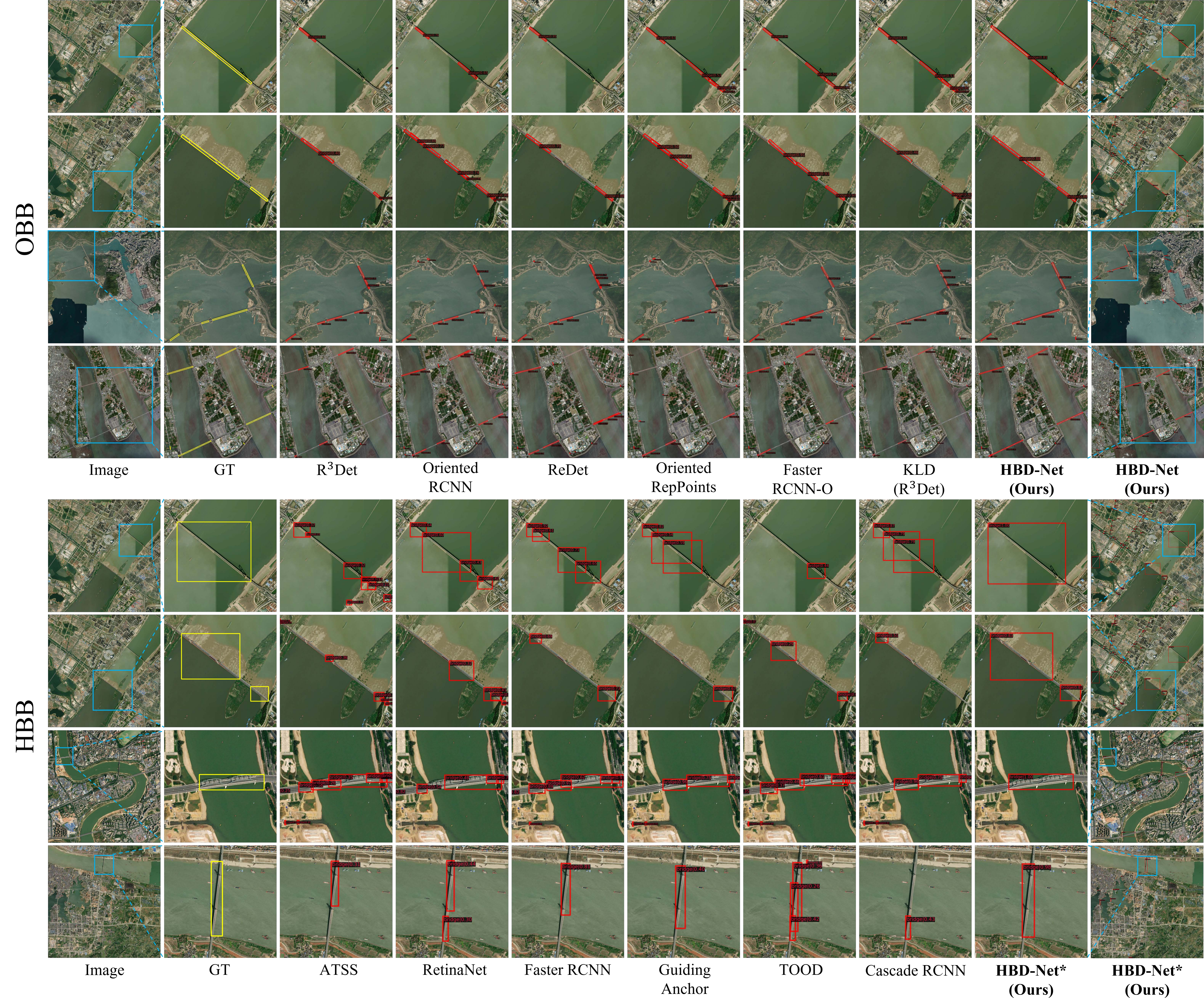}\\
           \vspace{-8pt}
	\caption{The visualization results of OBB and HBB tasks on the GLH-Bridge dataset using the HBD-Net and comparison object detection methods. * indicates using the HBD-Net without the proposed SSRW strategy.}
	\label{fig:obbvis}
\end{figure*}

\subsubsection{Evaluation Metrics}

We establish a benchmark on the GLH-Bridge dataset for two types of object detection tasks: \textbf{OBB} detection and \textbf{HBB} detection. The Average Precision (AP) is adopted as the main evaluation metric in this study (by the IoU computation for the True Positive (TP), False Positive (FP), and False Negative (FN)). We adopt the PASCAL VOC 07 metric\cite{everingham2010pascal} to calculate the mean Average Precision (mAP). In the MS-COCO dataset \cite{lin2014microsoft}, the pixel area of the ground-truth boxes is used to determine \emph{small}, \emph{medium}, and \emph{large} scales to calculate the corresponding AP values, which has been widely used to assess various detection algorithms. However, it is important to note that bridges, despite varying significantly in lengths and aspect ratios, may appear to possess the same area in VHR images. Dividing bridges solely based on area may prove inadequate to accurately reflect the detection difficulty and overlooks the influence of image size constraints on the detection algorithm. 

In light of the aforementioned limitation, we propose new evaluation metrics according to the length of the longer side of the ground-truth boxes. Specifically, we define a set of pixel intervals as $\left\{ (0,50], (50,200], (200,800], (800,16384] \right\}$ to categorize bridges based on their lengths, classifying them as \textit{short}, \textit{middle}, \textit{large}, and \textit{huge}. The corresponding APs are denoted as AP$_{\mathrm{sh}}$, AP$_{\mathrm{md}}$, AP$_{\mathrm{lg}}$, and AP$_{\mathrm{hg}}$, respectively. It is important to note that the detection of huge bridges can often be a challenging task, as they may not be effectively and completely captured within a single sliding window when employing traditional cropping strategies.

\subsubsection{Implementation Details}  The algorithms employed in our experiments are from two open-source pytorch-based algorithm libraries, MMRotate\cite{zhou2022mmrotate} and MMDetection\cite{chen2019mmdetection}. These libraries integrate various state-of-the-art object detection algorithms, along with their corresponding backbone networks, feature extractors, and detectors. They enable the reproduction of the original accuracies of the respective algorithms within a unified algorithm framework, ensuring fairness. Hence, these two algorithm libraries were chosen for the benchmarks for our experiments.

Experiments are performed on a server with 1 Tesla V100 GPU and 16GB memory. The backbone networks are initialized with models pre-trained on ImageNet\cite{deng2009imagenet}. We adopt the ``2×" training schedule in MMRotate and MMDetection. The SGD optimizer is employed with a learning rate of 0.005, momentum of 0.9, and weight decay of 0.0001. When performing feature fusion among the detectors of the proposed SDFF, the learning rate is set to 0.001. A linear warm-up strategy is applied for the initial 500 iterations, with a rate of 1.0/3. As for the algorithms used to establish benchmark results, the batch size is set to 4. 

In the case of the HBD-Net utilized in this study, the batch size is set to 1 during training, and the learning rate is adjusted accordingly. The image processing strategies for training and testing follow the description in \secref{sec:method}, the downsampling ratio $\sigma$ is set to 2.0. When training the HBD-Net, we use a label filtering strategy to divide original labels into $n$ groups to calculate loss with the outputs of $n$ layers. To the $n$-th label group, the filtering threshold is represented as $[min_n,max_n)$, and the $min_n$ is set to $ 15\times 2^{(n-1)}$ pixels and the $max_n$ is set to $1448$ pixels (\ie, $1024\times \sqrt{2}$ pixels) considering the size of the cropping window. In all experiments, random flipping was used as the only data augmentation technique.

\subsubsection{Mainstream Methods}

To assess the efficacy of the HBD-Net, we conduct a comparative evaluation against 18 advanced object detection methods. For the OBB task, we choose two-stage approaches such as Faster R-CNN-O \cite{xia2018dota}, RoI Transformer \cite{ding2019learning}, Oriented R-CNN \cite{xie2021oriented}, and ReDet \cite{han2021redet}; one-stage approaches including FCOS-O \cite{tian2019fcos}, R$^3$Det \cite{yang2021r3det}, KLD \cite{yang2021learning}, and Oriented RepPoints \cite{li2022oriented}; and methods for object detection in large-size images like CGL \cite{chen2023coupled}. Oriented R-CNN is chosen as the baseline method for the proposed HBD-Net and CGL. For the HBB task, we choose RetinaNet \cite{lin2017focal}, Faster R-CNN\cite{ren2015faster}, FCOS\cite{tian2019fcos}, TOOD \cite{feng2021tood}, Cascade R-CNN \cite{cai2018cascade}, ATSS \cite{zhang2020bridging}, GuidingAnchor \cite{wang2019region}; and methods designed for large-size images like CGL \cite{chen2023coupled}, GLSAN \cite{deng2020global}, and SAHI \cite{akyon2022sahi}. Faster R-CNN is chosen as the baseline method for CGL, GLSAN, SAHI, and HBD-Net on the HBB task. It should be noted that the SSRW strategy is not used when training the proposed HBD-Net on the HBB task.

In the case of CGL, GLSAN, and SAHI, we adopt their default strategies to process large-size images. For SAHI, we set the patch size to 1024$\times$1024 pixels, with a 200-pixel overlap if necessary, and combine three strategies: slicing-aided hyper inference, full image inference (FI), and an overlapping patches-based cropping strategy (PO). For the GLSAN, we adopt its default configuration for training and utilize its SelfAdaptiveCrop approach for testing with a crop size of 1024$\times$1024 pixels. For the other object detection methods, the original images are processed using an overlapping patches-based cropping strategy for training and testing. The cropping settings for training and testing are consistent, with a cropping window size of 1024$\times$1024 pixels and a 200-pixel overlap.

\subsubsection{Results and Analysis}
\begin{table*}[tb!] 
    \centering
    \caption{Accuracy ($\%$) of ablation studies on the impact of different strategies used in the proposed SDFF architecture on the OBB task on GLH-Bridge. ``DIP" denotes that using the proposed dynamic image pyramid. ``Unified detector" denotes that all layers in the SDFF use a unified object detector. ``Separate detector" denotes that each layer in the SDFF uses a separate detector. ``Distillation" denotes using cross-scale-transfer distillation in the training phase of the SDFF.
    }
    \resizebox{1.0\textwidth}{!}{
    \renewcommand\arraystretch{1.2}
    \setlength\tabcolsep{2.2mm}
    \begin{tabular}{c l c c | c c c|c c c c}
    \hline
        Structure & Setting & Distillation & IFF &$\mathrm{mAP}$ & $\mathrm{AP_{50}}$ & $\mathrm{AP_{75}}$ & $\mathrm{AP_{sh}}$ & $\mathrm{AP_{md}}$ & $\mathrm{AP_{lg}}$ & $\mathrm{AP_{hg}}$  \\ \hline
        \multirow{4}{*}{DIP} & Unified detector & $\times$  & $\times$
        & 34.61 & 69.95 & 26.71 & 32.97    & 38.15     & 30.79   & 12.36  \\ 
         & Separate detector & $\times$ & $\times$
        & 34.65 & 69.93 & 26.83 & 33.02    & 38.06     & 30.47   & 14.25 \\
         & Separate detector    & \checkmark & $\times$ 
        & 34.87 & 70.18 & 27.34 & 33.11    & 38.23     & 31.22   & 15.76   \\
         & Separate detector & \checkmark & \checkmark & \textbf{35.08} & \textbf{70.58} & \textbf{28.06} & \textbf{33.27}    & \textbf{38.51}     & \textbf{32.11}   & \textbf{17.30} \\ \hline
    \end{tabular}
    }
    \label{table_ablation2}
\end{table*}

The benchmark and experimental results for OBB and HBB tasks on GLH-Bridge are presented in Table \ref{tab:obbres}. 

For the OBB task, the experimental results demonstrate that the HBD-Net achieves the best performance on the benchmark of GLH-Bridge, with an $\mathrm{mAP}$ score of 35.35$\%$. It achieves an accuracy of 28.69$\%$ on the $\mathrm{AP_{75}}$ metric, underscoring the efficacy of our approach in accurately detecting rotated bridges. Furthermore, our method achieves the best performance, 33.47$\%$ and 20.61$\%$ in the $\mathrm{AP_{lg}}$ and $\mathrm{AP_{hg}}$  metrics, respectively. This highlights the HBD-Net's effectiveness in handling the detection of \textbf{large bridges} that may exceed the typical cropping size, particularly for instances with a length exceeding 800 pixels. Additionally, our method also shows benefits in detecting small objects, which constitute a significant portion of the dataset.

\begin{table*}[tb!]
    \centering
    \caption{Accuracy ($\%$) of ablation studies on the impact of the SDFF architecture and the SSRW strategy on the OBB task of GLH-Bridge. $\mathrm{SSRW_l}$ denotes using the SSRW strategy only on the baseline detector or the detector of the bottom layer of SDFF. $\mathrm{SSRW_g}$ denotes using the SSRW strategy only on the detectors of the layers except for the bottom layer of SDFF.}
    \resizebox{1.0\textwidth}{!}{
    \renewcommand\arraystretch{1.2}
    \setlength\tabcolsep{3.0mm}
    \begin{tabular}{c c c| c c c |c c c c}
    \hline
        $\mathrm{SSRW_l}$ & $\mathrm{SDFF}$ & $\mathrm{SSRW_g}$ &$\mathrm{mAP}$ & $\mathrm{AP_{50}}$ & $\mathrm{AP_{75}}$ & $\mathrm{AP_{sh}}$ & $\mathrm{AP_{md}}$ & $\mathrm{AP_{lg}}$ & $\mathrm{AP_{hg}}$  \\ \hline
          $\times$ & $\times$ & $\times$  & 34.16          & 69.87          & 26.29          & 32.83          & 37.74          & 29.30          & 5.68    \\ 
         \checkmark & $\times$ & $\times$  & 
        35.20  & 70.94  & 27.85  & 33.34 & 38.11 & 30.67 & 8.98   \\
         $\times$ & \checkmark & $\times$ &
         35.08 & 70.58& 28.06 & 33.27 & 38.51& 32.11& 17.30 \\
          \checkmark & \checkmark & $\times$ & 
        35.12 & 71.28 & 28.37 & \textbf{33.44} & 38.82 & 32.76 & 18.32   \\ 
         \checkmark  & \checkmark & \checkmark &
          \textbf{35.35} & \textbf{71.69} & \textbf{28.69} & 33.38   & \textbf{38.93} & \textbf{33.47} & \textbf{20.61}  \\ \hline
    \end{tabular}
    }
    \label{table_ablation} 
\end{table*}

For the HBB task, we consider that the aspect ratio of the horizontal box is determined by both the orientation of bridges and their true aspect ratios. Therefore, it does not accurately reflect whether the bridges are elongated in shape. As a result, we do not incorporate the SSRW strategy for the HBB task, only utilizing the proposed SDFF architecture as the employed approach. Under this setting, the HBD-Net also achieves a remarkable performance of 34.49$\%$ $\mathrm{mAP}$. Furthermore, in comparison to general object detection methods, the HBD-Net showcases outstanding performance in detecting large bridges. It obtains 35.21$\%$ and 35.59$\%$ in the $\mathrm{AP_{lg}}$ and $\mathrm{AP_{hg}}$ metrics, respectively. 

Additionally, for the methods designed for object detection in large-size images, although SAHI achieves fine small object detection by resizing overlapping patches, its upsampling technique provides limited benefits for VHR RSIs. CGL employs a fixed downsampling strategy, which results in information loss and suboptimal performance in $\mathrm{AP_{hg}}$. GLSAN performs prediction on the downsampled original image and selects sub-blocks for detailed detection through clustering of the predicted results. However, it tends to miss scattered small bridges, and is still hard to comprehensively detect large bridges. 

\begin{table}[tb!]
    \centering
    \caption{Accuracy ($\%$) of ablation studies on the OBB task on DOTA-v1.0. $\mathrm{SSRW}$ denotes using the SSRW strategy only on the detector of the bottom layer of SDFF.}
    \resizebox{0.48\textwidth}{!}{
    \renewcommand\arraystretch{1.2}
    \setlength\tabcolsep{4.4pt}
    \begin{tabular}{c c|c c c|c}
    \hline
     $\mathrm{SSRW}$  & $\mathrm{SDFF}$  & $\mathrm{mAP}$   & $\mathrm{AP_{50}}$  & $\mathrm{AP_{75}}$  & $ \mathrm{AP_{BR}}$  \\ \hline
     $\times$  &  $\times$  & 44.92 & 75.80 & 45.45 & 54.52   \\ 
      \checkmark   &  $\times$  & 45.63 & 76.25 & 45.98 & 55.48  \\
       $\times$ &  \checkmark  & 45.92 & 76.53 & 46.32 & 55.78   \\
     \checkmark & \checkmark   & \textbf{46.11} & \textbf{76.95} & \textbf{46.76} & \textbf{56.02 }\\ \hline
    \end{tabular}
    }
    \label{table_ablation_dota}   
\end{table}
    
In conclusion, the experimental results from both the OBB and HBB tasks demonstrate the effectiveness of the HBD-Net in a general sense. It is capable of adapting to the characteristics of both horizontal and oriented bounding boxes, and the visual results are shown in \figref{fig:obbvis}. Additionally, our HBD-Net is independent of the specific object detection methods. Therefore, it can seamlessly accommodate a wide range of advanced one-stage or two-stage object detectors within the proposed SDFF without encountering specific limitations. This observation highlights the versatility and applicability of the proposed approach in this study.

\subsection{Component Analysis}

We conduct ablation experiments on the GLH-Bridge dataset to evaluate the impact of two key components in our proposed HBD-Net (\ie, the SDFF architecture and the SSRW strategy).

\subsubsection{Effectiveness of the SDFF}
\label{sec:DIPFF}

As shown in Table \ref{table_ablation2}, we explore the effectiveness of the detector utilization strategy and IFF used in the proposed SDFF architecture. Our proposed SDFF without cross-scale-transfer distillation and IFF demonstrates a significant enhancement in accurately detecting large bridges, with a notable 8.57$\%$ improvement in $\mathrm{AP_{hg}}$ metric compared to the baseline. When considering whether each layer in the SDFF employs an individual detector or if all layers share a detector, the former slightly outperforms the latter. When we incorporate a cross-scale-transfer distillation strategy into the process of training the SDFF, the accuracy can be improved, resulting in an additional improvement of 3.4$\%$ improvements in $\mathrm{AP_{hg}}$ metric. Furthermore, through the integration of the IFF module, the higher layer can benefit from the finer details provided by the lower layer, leading to the improved final performance in terms of the $\mathrm{AP_{75}}$ and $\mathrm{AP_{hg}}$ metrics, which reach 28.06$\%$ and 17.30$\%$, respectively.

\subsubsection{Effectiveness of the SSRW strategy}

As our proposed HBD-Net utilizes respective detectors in the proposed SDFF, as shown in Table \ref{table_ablation}, we examine the effectiveness of the SSRW strategy when applied to these detectors individually. It can be observed that the proposed SSRW strategy enhances the regression accuracy of the detector when it is applied solely to the detector of the bottom layer, it results in 1.37$\%$ and 3.30$\%$ improvements in $\mathrm{AP_{lg}}$ and $\mathrm{AP_{hg}}$ metrics, respectively, compared to the baseline. Furthermore, with the incorporation of the SDFF architecture, we extend the application of the SSRW strategy to detectors corresponding to the higher layers of the pyramid, leading to a further improvement of 2.29$\%$ in $\mathrm{AP_{hg}}$ metric. Given the typically larger aspect ratios of large bridges, the above experiments demonstrate the effectiveness of our proposed SSRW strategy in directing the network's focus toward bridges with larger aspect ratios, thereby improving detection accuracy. Finally, in addition to the overall improvement in all metrics, the HBD-Net achieves a significant improvement of 2.40$\%$, 4.17$\%$ and 14.93$\%$ in $\mathrm{AP_{75}}$, $\mathrm{AP_{lg}}$ and $\mathrm{AP_{hg}}$ metrics, respectively, compared to the baseline. This study affirms the effectiveness of the proposed method in enhancing bridge detection performance in large-size images, especially concerning the detection of large bridges in their entirety. It is important to note that, with the implementation of the SSRW strategy for higher-layer detectors within the SDFF, a decrease in the $\mathrm{AP_{sh}}$ metric was observed. This decrease is attributed to a decrease in the proportion of small-scale labels at the higher layers resulting from the label filtering. As a result, SSRW's role in maintaining the balance of loss between small and large objects is diminished, aligning with its intended design principles.

Moreover, to comprehensively evaluate the effectiveness of our method, we further conducted ablation experiments on the DOTA-v1.0 dataset\cite{xia2018dota}. These experiments demonstrate how our designed modules progressively enhance the performance step by step. As shown in Table \ref{table_ablation_dota}, our proposed SSRW strategy and SDFF architecture result in 0.96$\%$ and 1.26$\%$ improvement respectively in $\mathrm{AP_{BR}}$ metric. Our HBD-Net achieves 46.11$\%$ $\mathrm{mAP}$ and 56.02$\%$ $\mathrm{AP_{BR}}$ based on the baseline. These results highlight the capability of our proposed HBD-Net to enhance the performance of existing state-of-the-art object detection methods.

\subsection{Cross-Dataset Generalization Experiments}
\subsubsection{Datasets}
We choose two public datasets (DOTA-v1.0 and DIOR-R) for the cross-dataset generalization experiments. These datasets are chosen based on their large-scale and diverse data characteristics, making them fundamental benchmarks in the field of remote sensing object detection. The details are as follows.

\textbf{DOTA-v1.0} \cite{xia2018dota}: DOTA-v1.0 is a large-scale dataset for object detection in aerial images. Its training and validation sets contain a total of 2,541 bridges in 288 images. 

\textbf{DIOR-R} \cite{cheng2022anchor}: DIOR-R is a large aerial images dataset and has various spatial resolutions, containing 4,000 bridges among 1,576 images with OBB annotation. For the DIOR-R dataset, the provided training, validation, and testing sets are utilized for the cross-dataset generalization experiments.

\subsubsection{Experimental Setting}
Cross-dataset generalization analysis is an important evaluation method for assessing the generalization performance of a dataset. We conduct cross-dataset generalization experiments using the bridge subset of the DOTA-v1.0 dataset\cite{xia2018dota} and the DIOR-R dataset \cite{cheng2022anchor}. For the DOTA-v1.0 dataset, we extract the bridge subset from the official training and validation sets for training purposes. The inference is performed on the official unlabeled test set using the standard format. Finally, the test results are uploaded to the official server to obtain accuracy. For the DIOR-R dataset, we select the bridge subset within the provided training and validation sets for training and evaluate the official test set. 

We employ Oriented R-CNN\cite{xie2021oriented} as the algorithm for training and testing. We train models on these three datasets respectively and conduct cross-dataset evaluation. The training settings for DOTA-v1.0 and DIOR-R are kept consistent with the original papers, both with the ``1×" training schedule \cite{zhou2022mmrotate}. For our constructed GLH-Bridge dataset, we utilize the training set to train our models while maintaining consistent training settings with the benchmark baseline. To ensure image size compatibility, we implement a cropping strategy with window sizes of 1024$\times$1024 pixels for DOTA-v1.0 and 800$\times$800 pixels for DIOR-R, along with a 200-pixel overlap. The evaluation of cross-dataset generalization experiments is conducted based on the $AP$ metric. 

\begin{table}[!htb]
    \centering
    \caption{Accuracy ($\%$) on cross-dataset generalization experiments. }
    \resizebox{0.48\textwidth}{!}{
    \renewcommand\arraystretch{1.2}
    \setlength\tabcolsep{1.6pt}
    \begin{tabular}{l|c|c|c}
    \hline
     \diagbox[width=7em]{Test on}{Train on} & GLH-Bridge & DOTA-v1.0 & DIOR-R \\ \hline
     
    GLH-Bridge & \textbf{34.16}   & 15.78 & 15.01 \\ 
    DOTA-v1.0 & \textbf{49.55}   & 45.76   & 18.46  \\ 
    DIOR-R & \textbf{20.74}  & 12.14    & 19.88  \\ \hline
    \end{tabular}
    }
    \label{tab:CrossDatasetExp}
    \end{table}

\begin{figure}[!htb]
  \centering
  \includegraphics[width=\columnwidth]
  {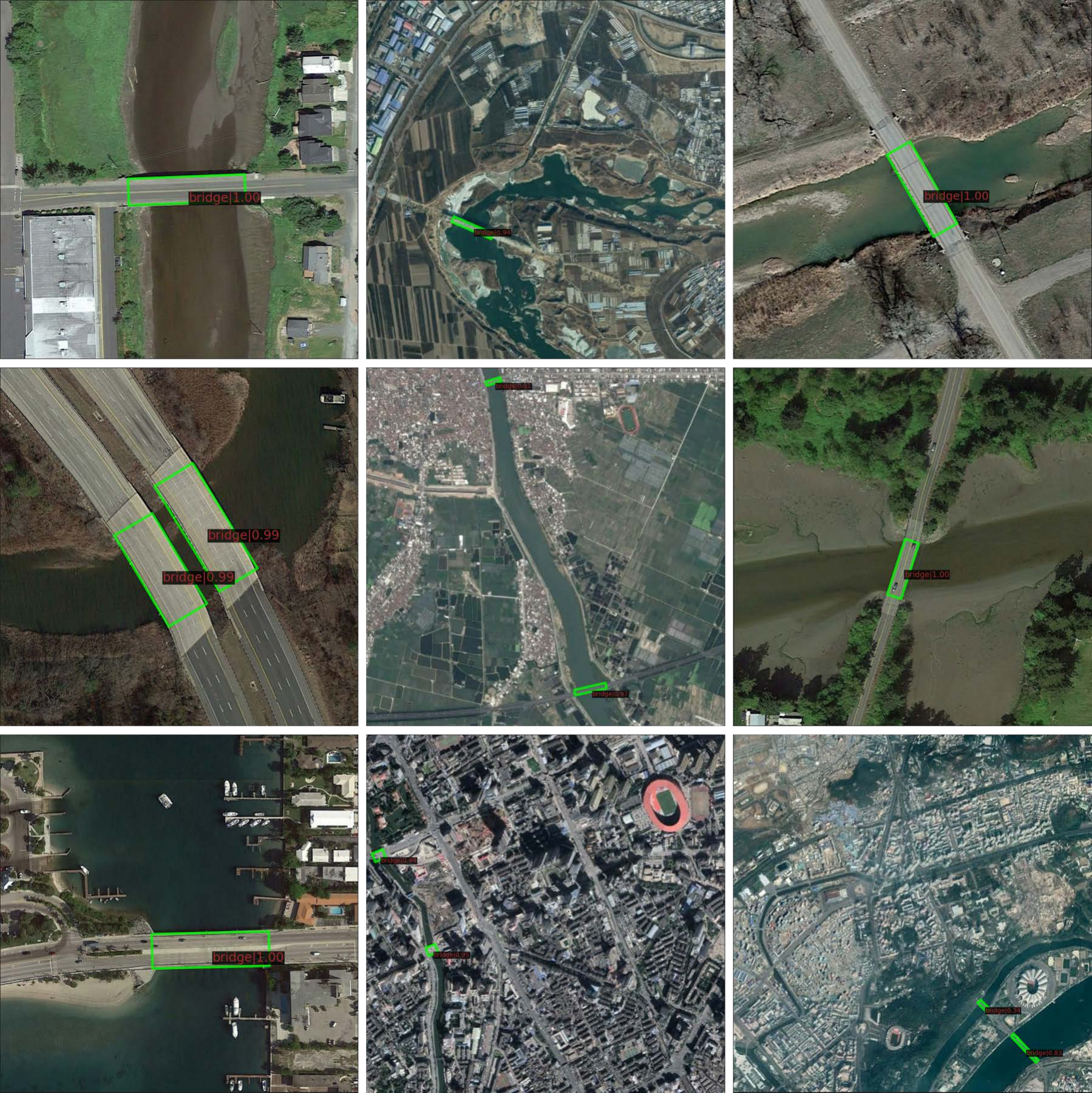}\\
  \caption{The visualized prediction results of the DIOR-R dataset by the model trained on the GLH-Bridge dataset.}
  \label{fig:DIOR-R_vis}
\end{figure}

\begin{figure}[!htb]
  \centering
  \includegraphics[width=\columnwidth]
  {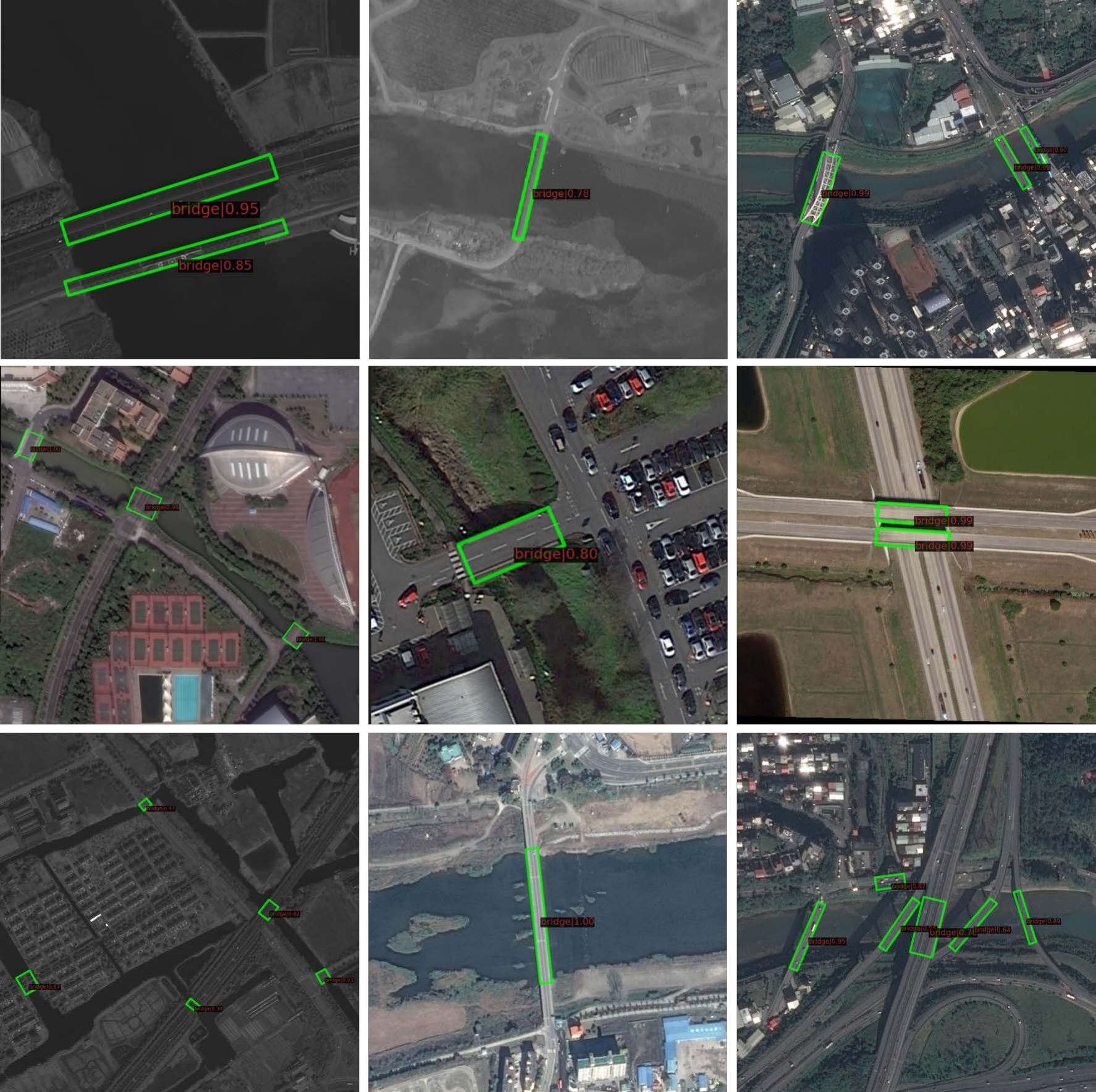}\\
  \caption{The visualized prediction results of the DOTA-v1.0 dataset by the model trained on the GLH-Bridge dataset.}
  \label{fig:dota_vis}
\end{figure}

\subsubsection{Results and Analysis}
The experimental results are shown in Table \ref{tab:CrossDatasetExp}, which show that GLH-Bridge has achieved excellent zero-shot generalization results on two mainstream benchmarks. Specifically, it has achieved a performance improvement of 3.79$\%$ on the DOTA-v1.0 dataset and 0.86 $\%$ on the DIOR-R dataset. These results indicate that the ability of the GLH-Bridge dataset to provide a more comprehensive and accurate representation of bridge characteristics within the domain of the perspective of remote sensing imagery.

The visual results of the DIOR-R dataset generated by the model trained on the GLH-Bridge dataset are shown in \figref{fig:DIOR-R_vis}. It can be observed that despite the significant variation in image resolution within the DIOR-R dataset (ranging from 0.5m to 30m), the model trained on GLH-Bridge exhibits the capability to identify bridges in low-resolution images. Additionally, the DIOR-R dataset contains bridges with diverse color tones and extreme aspect ratios. Despite differences in satellite sources between these images and those in the GLH-Bridge dataset, the model trained on GLH-Bridge demonstrates strong generalization ability by successfully detecting bridges in backgrounds with high interference and images with lower resolutions.

The visualized prediction results of the DOTA-v1.0 dataset by the model trained on the GLH-Bridge dataset are shown in \figref{fig:dota_vis}. It can be observed that despite the inclusion of panchromatic remote sensing images in addition to RGB images in the DOTA-v1.0 dataset, the proposed model trained on the GLH-Bridge dataset is still able to accurately identify bridges. This demonstrates that the GLH-Bridge dataset can capture the core features of bridges in remote sensing images, which are invariant to color. Moreover, the trained model achieves good performance in identifying small bridges in the DOTA-v1.0 dataset, which proves that the GLH-Bridge dataset has meticulous and high-quality annotations.

\section{Conclusion}\label{sec:conclusion}
In this paper, we propose a large-scale dataset named GLH-Bridge for holistic bridge detection in large-size VHR RSIs. The proposed dataset consists of 6,000 VHR RSIs, with image sizes ranging from 2,048 $\times$ 2,048 to 16,384 $\times$ 16,384 pixels, and contains 59,737 bridges spanning diverse backgrounds with OBB and HBB annotation. The large image size, the large sample volume, and the diversity of object scale and background type make GLH-Bridge a valuable dataset, which has the premise to promote one new challenging but meaningful task: holistic bridge detection in large-size VHR RSIs. Furthermore, we present the HBD-Net, a cost-effective solution tailored for holistic bridge detection in large-size images. Based on the proposed GLH-Bridge dataset, we establish a benchmark and provide empirical validation of the effectiveness of the proposed HBD-Net. In future work, we will continue to enrich the GLH-Bridge dataset in terms of its sample volume and sub-category annotation. Additionally, we aim to generalize the proposed HBD-Net to address multi-class object detection in large-size images.

\ifCLASSOPTIONcaptionsoff
  \newpage
\fi

\bibliographystyle{IEEEtran}
\bibliography{arxiv}

\end{document}